\documentclass[pdflatex,sn-basic]{sn-jnl}

\usepackage{multirow}
\usepackage{pifont}
\usepackage{wrapfig}
\usepackage{caption}
\usepackage{url}
\usepackage{subcaption}
\usepackage{booktabs}

\jyear{2023}%

\begin{document}

\title{Boosting Deep Neural Networks with Geometrical Prior Knowledge: A Survey}


\author*[1,2]{\fnm{Matthias} \sur{Rath}}\email{Matthias.Rath@de.bosch.com}

\author*[1,2]{\fnm{Alexandru Paul} \sur{Condurache}}\email{AlexandruPaul.Condurache@de.bosch.com}

\affil[1]{\orgdiv{Cross Domain Computing Solutions}, \orgname{Robert Bosch GmbH}, \orgaddress{\city{Stuttgart}, \country{Germany}}}

\affil[2]{\orgdiv{Institute for Signal Processing}, \orgname{University of L\"ubeck}, \orgaddress{\city{L\"ubeck}, \country{Germany}}}

\abstract{Deep Neural Networks achieve state-of-the-art results in many different problem settings by exploiting vast amounts of training data. However, collecting, storing and - in the case of supervised learning - labelling the data is expensive and time-consuming. Additionally, assessing the networks' generalization abilities or predicting how the inferred output changes under input transformations is complicated since the networks are usually treated as a black box. Both of these problems can be mitigated by incorporating prior knowledge into the neural network. 
One promising approach, inspired by the success of convolutional neural networks in computer vision tasks, is to incorporate knowledge about symmetric geometrical transformations of the problem to solve that affect the output in a predictable way. This promises an increased data efficiency and more interpretable network outputs. In this survey, we try to give a concise overview about different approaches that incorporate geometrical prior knowledge into neural networks. Additionally, we connect those methods to 3D object detection for autonomous driving, where we expect promising results when applying those methods.
}




\maketitle

\section{Introduction}
Deep Neural Networks (DNNs) achieve state-of-the-art results on various tasks such as speech recognition, object detection, text generation or machine translation \citep{DeepLearning}. Usually, DNNs are trained on a large amount of training data in order to generalize to similar, but unseen test data. However, gathering labeled data, which is needed for supervised learning methods, is both labor-intensive and time-consuming. Thus, it is desirable to increase the data efficiency to obtain a good performance even when available data is limited. 

Furthermore, DNNs suffer from some key disadvantages, especially when applied in fields with strict safety requirements such as robotics, autonomous driving or medical imaging. In these fields, DNNs need to be robust and explainable to allow assessing their behavior in safety-critical cases. This contradicts the current approach of treating DNNs as a black box and training them in an end-to-end fashion.
Finally, the solution space of all possible DNN parameters is high-dimensional which impedes finding the optimal solution to the learning problem. Thus, it makes sense to restrict the solution space using problem-dependent, reasonable constraints.

Consequently, current research tries to combine expert knowledge -- e.g. the knowledge utilized to design classical pattern recognition systems -- with the architectures and optimization methods of DNNs. This is often called a \textbf{hybrid approach} aiming to combine the best of both worlds: state-of-the-art results by data-driven optimization while obtaining improved performance from limited data. Additionally, an explainable, robust system behavior is desired.

Expert knowledge can be expressed in many different ways. A simple example is knowledge about the expected size of different objects in images for object detection, used for the \textit{anchor boxes} in object detectors. A more sophisticated example is the exact definition of a filter sequence used to detect those objects. However, this kind of prior knowledge is hard to formalize which makes it difficult to include it into a DNN in a principled manner. Another possibility is to model a prior distribution over the output of a neural network in order to predict a posterior distribution utilizing a Bayesian neural network.  

For many tasks, certain transformations of the input can be determined that affect the desired output in a predictable way -- or not at all. Both, knowledge about physical transformations, e.g. when a camera is moved to a novel viewpoint \citep{CoorsNoVA}, or knowledge about certain transformations frequently occurring in the input data, e.g. rotated and translated patterns, can be leveraged. This type of prior knowledge is called \textbf{geometrical prior knowledge}.

In general, geometrical prior knowledge can be applied by forcing the DNN's output or learned features to be \textbf{in- or equivariant}. A function is \textbf{invariant} with respect to (w.r.t.) a transformation, if its output does not change under transformations of the input. Comparably, the function is \textbf{equivariant}, if input transformations induce predictable transformations in the output space. For example, an image classifier should classify a dog correctly independent of its orientation. Therefore, it needs to learn a representation that does not change when the input is rotated, i.e. it should be invariant w.r.t. rotations. If we also aim to predict the dog's orientation, an equivariant representation that is guaranteed to rotate under input rotations is beneficial.

A common approach to obtain a DNN that is robust to input transformations is \textbf{data augmentation}, where the input is randomly transformed during training. While data augmentation is flexible and straightforward to implement, the DNN only learns to approximate the desired in- or equivariance properties. The prior knowledge is not incorporated in a mathematically guaranteed way. 

A well-known example of incorporating geometrical prior knowledge to DNNs in a mathematically guaranteed way are \textbf{convolutional} neural networks (CNNs) which share a learnable kernel among all input locations. This procedure is called \textit{translational weight tying} and allows to reduce the parameter count of DNNs while also facilitating the DNN to recognize patterns independent of their location in the input. Therefore, CNNs are \textbf{equivariant} to \textbf{translations}. The success of CNNs in computer vision tasks confirms that utilizing geometrical prior knowledge is an important \textit{inductive bias} for DNNs. In general, the concept of CNNs can be generalized to enforce equivariance towards more complicated transformations than translations.

In this contribution, we present methods that integrate geometric prior knowledge to DNNs such that their representations are 
in- or equivariant. First, we review the mathematical concepts underlying in- and equivariant representations. We then provide an overview of different approaches which allow to enforce those properties. We group related work into different sub-fields and summarize the contributions along those categories. Afterwards, we give an overview over common datasets and benchmarks which are used to compare the different presented algorithms. We provide a brief glimpse how those methods can be applied to DNNs used in the autonomous driving context. Finally, we summarize our review paper and give a short outlook on open challenges and future work. 

\textbf{Remark:} In this paper, we mainly focus on work related to perception for autonomous driving, i.e. computer vision and processing 3D sensor data. We mention some work incorporating geometrical prior knowledge to other domains, but do not claim any completeness. We do not present any new results but hope to give a broad overview over geometrical prior knowledge applied to DNNs. Thereby, we hope to provide an easy entry into this interesting field for novel researchers and the possibility to compare different approaches for experienced researchers. 

\section{Preliminaries}
In this section, we briefly introduce the core concepts needed to understand the work presented in our survey. Mainly, we discuss the mathematical concept of groups, in- and equivariance, group representation theory and steerable filters.
\subsection{Group Theory}
Groups are a mathematical abstraction that can be used to model invertible geometrical transformations. A group $G$ consists of a set of elements and a group operation, which combines two elements of the group to form a third $ab=c$ with $a,b,c \in G$. It fulfills the axioms of closure, associativity, identity and invertibility.

Group theory is the basis to mathematically describe geometrical symmetries. It is used to formally define in- and equivariance w.r.t. transformations $g \in G$. A function $f$ is equivariant w.r.t. a transformation group $G$, if there exists an explicit relationship between transformations $\mathcal{T}_g^{\mathcal{X}}$ of the function's input and the corresponding transformation $\mathcal{T}_g^{\mathcal{Y}}$ of its output. Here, the transformation acts on the input vector space $X\in\mathbb{R}^n$ via the left group action $g \times \mathbb{R}^n \rightarrow \mathbb{R}^n$, $(g,x) \mapsto\mathcal{T}_g^{\mathcal{X}}[x]$ with $g \in G$.
\begin{equation}
f(\mathcal{T}_g^{\mathcal{X}}[x]) = \mathcal{T}_g^{\mathcal{Y}}[f(x)] \; \forall \; g\in G, x \in X \text{.}
\end{equation}
Similarly, the left group action $g^\prime \times \mathbb{R}^m \rightarrow \mathbb{R}^m$, $(g^\prime,f(x)) \mapsto\mathcal{T}_g^{\mathcal{Y}}[f(x)]$ with $g^\prime \in G^\prime$ describes the induced transformation in the output space $f(x) \in \mathbb{R}^m$. It is important to note, that $\mathcal{T}_g^{\mathcal{X}}$ and $\mathcal{T}_g^{\mathcal{Y}}$ do not need to be the same transformation. For example, rotating the input might induce a shift in the output space.

Usually, a simplified notation that directly uses the group element $g \in G$ as a drop-in replacement for the left group action is used for the definition of equivariance
\begin{equation}
f(gx) = g^\prime f(x) \; \forall \; g \in G, \; x \in X \text{.}
\end{equation}

Invariance is a special case of equivariance, where the transformation in output space is the identity, i.e. the output is unaffected by input transformations $g \in G$
\begin{equation}
f(gx) = f(x) \quad \; \forall \; g \in G, \; x \in X \text{.}
\end{equation}

As mentioned, a common example of an equivariant function $f$ are convolutional layers, which introduce translation equivariance. 
Depending on the properties of the task to solve, equivariance is more suitable than invariance as it preserves information about the symmetry group. For example, a 3D object detector needs to be equivariant to rotations since one of the tasks is to estimate the detected objects' rotation angles. Invariance would destroy this information in the feature space, which would be beneficial for the pose-independent classification of those objects.

\subsection{Group Representation Theory}
The mathematical field of Group Representation Theory investigates, how a group action $\mathcal{T}_g \in G$ acts linearly on a $n$-dimensional complex vector space $\mathbf{V}$. This is interesting in the context of DNNs because feature spaces can be modeled as vector spaces while symmetric input transformations can be modeled as groups. Hence, Group Representation Theory provides the mathematical backbone on how feature spaces of CNNs change under input transformations.

For matrix Lie groups, a finite-dimensional complex representation $\Pi$ of $G$ is a group homomorphism $\Pi : G \to \text{GL}(\text{dim}(\mathbf{V}))$, that maps the group to the general linear group $\text{GL}(\mathbb{C}^n)$, i.e. to the group of invertible $n\times n$ matrices representing the linear action of the group on the vector space \citep{HallGroupRepresentation}.

We define three common group representation types which are generally used for in- or equivariant representations in the DNN literature. The \textbf{trivial representation} maps all group elements to the identity matrix, i.e. the vector space $\mathbf{V}=\mathbb{C}^n$ is left invariant under group transformations. 

The \textbf{regular representation} is determined by the action of $G$ on itself. In this case, $\mathbf{V}=C^{n\times\|G\|}$ is a $\|G\|$-dimensional vector space which is permuted under group actions $g \in G$.

A \textbf{representation} is called \textbf{irreducible} (\textbf{irrep}), if its only invariant subspaces are the trivial subspaces $\mathbf{W}=\mathbf{V}$ and $\mathbf{W}=\{0\}$. A subspace $\mathbf{W}$ of $\mathbf{V}$ is called invariant, if $\Pi(g)w \in \mathbf{W}$ for all $w \in \mathbf{W}$ and all $g \in G$. Importantly, \textit{Maschke's theorem} states that every representation of a finite group consists of a \textit{direct sum} of irreducible representations. Consequently, finding all irreducible representations of a specific group is an interesting aspect of group representation theory. Furthermore, irreps provide the smallest possible representation that can be used to incorporate equivariance. Irreps canoften  even be used to achieve invariance to continuous groups.

For a more detailed discussion of group representation theory which forms the backbone of equivariant neural networks, we refer the interested reader to the recent publications by \cite{E2STCNNs} and \cite{EstevesSurvey} or to the text book by \cite{HallGroupRepresentation}.

\subsection{Steerable Filters}\label{sec:SteerableFilters}
Steerable filters \citep{Adelson} are filters whose arbitrary rotated version $f^\theta$ can be synthesized using a linear combination of a finite number of basis filters $\psi_i$ . In the two-dimensional case, rotation-steerable filters are defined as:
\begin{equation}
f^\theta(r,\phi) = \sum_{i=1}^{M}w_i(\theta)\psi_i(r,\phi)\text{,} 
\end{equation}
where $\theta$ is the rotation angle and $w_i(\theta)$ are called interpolation functions. Steerable filters are defined in the polar space $r=\sqrt{x^2+y^2}$, $\phi = \arctan(\|\frac{y}{x}\|)$. 
Steerable filters can be calculated in arbitrary rotated versions analytically without suffering from sampling artifacts. This is important for computer vision tasks, where multiple rotated versions of a filter are applied or learned frequently. 

The concept of steerable filters can also be generalized to leverage those advantages for arbitrary transformations $G$. The transformed steerable filters can again be computed in closed form for arbitrary transformations $h \in G$ via a sum of basis filters
\begin{equation}
f^h(g) = \sum_{i=1}^{M}w_i(h)\psi_i(g)\text{.} 
\end{equation}

\section{A Taxonomy of Geometrical Prior Knowledge for Deep Neural Networks}
\begin{figure}[h]
	\centering
	\includegraphics[width=\textwidth]{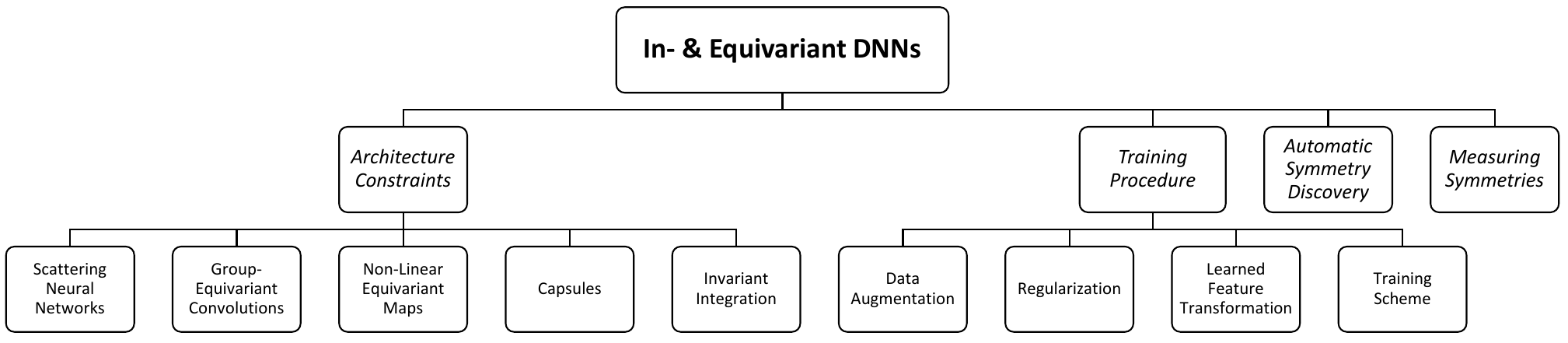}
	\caption{Taxonomy of methods leveraging Geometrical Prior Knowledge for DNNs.}
	\label{fig:topcis}
\end{figure}
In general, geometrical prior knowledge can be incorporated to DNNs in order to restrict the solution space of the learning algorithm. If the prior knowledge is carefully chosen, i.e. it is relevant for the task the network is trying to solve, this helps to raise the data efficiency of the learning process. Since the available amount of training data is finite incorporating geometrical prior knowledge leads to an improved overall performance of DNNs, especially for small datasets.

Geometrical prior knowledge can be enforced to DNNs in multiple ways. We provide a general Taxonomy of the in- and equivariant DNN literature in Figure \ref{fig:topcis}. Following this Taxonomy, we present recent research that achieves \textit{guaranteed equivariance} via \textbf{architecture restrictions} (Section \ref{sec:Guaranteed}), learns \textit{approximate equivariance} by adapting the \textbf{training procedure} (Section \ref{sec:Approximated}), automatically \textbf{discovers in- or equivariance} from data (Section \ref{sec:Discovered}) and \textbf{measures equivariance} properties (Section \ref{sec:Measured}).

\section{Architecture Constraints}\label{sec:Guaranteed}
In general, in- or equivariance can be incorporated into a DNN in a mathematically guaranteed way by restricting the architecture or the learnable filters of the DNN. \textbf{Convolutional Neural Networks} (CNNs) share a convolutional kernel among all positions of the input image or feature map \citep{Fukushima, LeCun}. Thereby, CNNs are translation equivariant, which is one of the key properties responsible for their success in various practical applications. However, standard convolutions are not equivariant to other transformations such as rotations. Thus, changing or generalizing the convolution to enable equivariance to arbitrary transformations is a promising approach to increase the data-efficiency of CNNs.

\subsection{Scattering Neural Networks}
In- or equivariance properties can be enforced to functions by using a well-defined map with fixed filters that yields the desirable properties.
\begin{table*}[t]
	\centering
	\resizebox{\columnwidth}{!}{
	\begin{tabular}{cccc}
		\toprule
		Reference & Invariance group & Classifier & Learnable \\ 
		\midrule
		\cite{Scattering} & $\mathbb{Z}^2$ & SVM & \ding{55} \\
		\cite{GenericScattering} & $\mathbb{Z}^2$ & Linear SVM &  \ding{55} \\
		\cite{DeepHybridNetworks, OyallonCompressingScattering, Scattering2} & $\mathbb{Z}^2$ & CNN & \ding{55} \\
		\cite{ScatteringHomotopyDictionary} & $\mathbb{Z}^2$ & Dictionary + MLP & \ding{55}\\ 
		\cite{Cotter15, Cotter2} & $\mathbb{Z}^2$ & CNN & \checkmark\\
		\cite{ParametricScattering} & $\mathbb{Z}^2$ & CNN & \checkmark\\
		\cite{RotoTranslationScattering} & $\mathbf{SE}(2)$ &  Gaussian SVM & \ding{55} \\
		\cite{Sifre} & $\mathbf{SE}(2) \rtimes S$ & PCA classifier  & \ding{55} \\
		\bottomrule 
	\end{tabular}
	}
	\caption{Overview of Scattering CNNs classified by the invariance group $G$, the classifier used on top of the scattering coefficients and if the scattering coefficients are learnable.}
	\label{tab:Scatter}
\end{table*}
\cite{Scattering} propose to use a hand-crafted \textbf{scattering transformation} that enforces invariance to the representations of a DNN. It consists of a convolution of the input signal $x$ with a family of wavelets $\psi$ followed by a modulus non-linearity and an averaging filter $\phi$. The first-order scattering coefficients $S^{1}(x)$ at rotation and scale $\lambda_1 = \{\theta_1, j_1\}$ are defined as:
\begin{equation}
S^{1}(x) = \|x\star\psi_{\lambda_1}\|\star\phi_J \text{,}
\end{equation}
where the family of wavelets is computed by rotating and dilating a complex mother-wavelet $\psi$ using $L$ rotations and $J$ scales.
Furthermore, multiple scattering transforms can be cascaded to obtain coefficients of order $m\geq0$
\begin{equation}
S^{m}x = \| \hdots \|\| x \star \psi_{\lambda_1}\| \star \psi_{\lambda_2}\| \hdots  \star \psi_{\lambda_m}\| \star\phi_{J} \text{.}
\end{equation}

\cite{Scattering} use complex Morlet wavelets to obtain the scattering coefficients and process the features with a Support Vector Machine (SVM) for classification. While the calculated scattering coefficients are invariant to translations and stable to small deformations, they are robust, but not invariant to other transformation groups such as rotations.

\cite{GenericScattering} compare a two-layer scattering network for image classification to a DNN of the same depth and conclude that the fixed scattering filters closely resemble the filters learned within the early layers of a CNN during training.

\cite{GroupInvScattering} generally shows that the scattering operator can be generalized to arbitrary transformations that can be represented as compact Lie groups $G$. A prove for the invariance of the scattering transformation to the action of $G$ is provided. Moreover, the application to the group of rotations and translations in 2D is proposed, which is conducted by \cite{RotoTranslationScattering}. 

\cite{Sifre} further advance this approach by cascading multiple scattering transformations with global space-scale averaging and invariant linear projectors to obtain a representation that is invariant to translation, rotation, scaling and stable to small deformations.  

\cite{DeepHybridNetworks,Scattering2} introduce a hybrid scattering approach: The lower layers of a CNN are replaced by fixed scattering transformations while the upper layers are learned convolutions. Thus, the lower layer features are translation-invariant and fixed while the higher layers learn more abstract features. The proposed method enables a performance boost when training on limited training data subsets compared to a fully learnable CNN. This demonstrates that the scattering transform does not lose any discriminative power while boosting the performance in the limited sample domain by incorporating geometrical prior knowledge.

The scattering transformation can also be used to compress the input images processed by a CNN to reduce the spatial resolution. While this reduces parameters and training time, it still captures most of the signal information needed for classification permitting state-of-the-art classification results \citep{OyallonCompressingScattering}.

\cite{ScatteringHomotopyDictionary} combine the scattering transformations with a single dictionary calculating a positive $\ell^1$ sparse code of the scattering coefficients. The dictionary is learned via homotopy iterative thresholding algorithms in combination with a classification MLP. This simplified CNN using only well-defined mathematical operators is able to outperform a deep CNN (AlexNet) on ImageNet. At the same time, the exact network properties can be investigated via the learned dictionary.

\cite{Cotter} visualize the patterns that caused a large activation in each channel of a hybrid scattering CNN \citep{DeepHybridNetworks}. The filters used by scattering networks are then compared to the learned filters of conventional CNNs. Therefore, a method called DeScatterNet is used which sequentially inverts the operations of the scattering layers. In comparison, the CNN filters are more general, i.e. they are able to detect more complex shapes. To resolve this discrepancy, the authors propose to replace the averaging filters of the original scattering transformation by a more general filter which preserves the spatial domain. This procedure allows the scattering network to detect more complex shapes like corners, crosses and curves. 

\cite{Cotter2} propose a learnable variant of the scattering transformation called locally invariant convolutional layer by multiplying the output of the scattering transformation with learnable weights. This layer is differentiable w.r.t. the learnable weights and its input which makes it suitable to use for end-to-end training via the backpropagation algorithm. The novel layer is applied at different depths of a CNN and works best, when it is used in an early, but not the first layer of the CNN. 

\cite{ParametricScattering} directly learn the parameters of the scattering transform's Mother wavelet via back-propagation instead of using a filterbank with hand-crafted wavelets. Thereby, the design of the wavelet family has relaxed constraints, granting problem-dependent flexibility. Similar to \cite{DeepHybridNetworks}, the scattering transform is used in the early layers of DNNs. The parametric, more flexible approach further improves the limited data performance of scattering-enhanced DNNs.

To summarize, the scattering transformation allows to incorporate symmetry properties to neural networks while guaranteeing stability to small deformations, which is a desirable property for many learning tasks. While early proposals use the transformation in combination with classifiers such as a SVM, more recent approaches include it into DNN architectures. One approach is to replace the early layers of a DNN with a fixed scattering filterbank, i.e., the scattering coefficients are used as the input for learnable CNN layers. It is also possible to directly learn the parameters of the Mother wavelet via backpropagation or introduce learnable parameters to the scattering layer itself while ensuring that it is differentiable w.r.t. its input, making it suitable for end-to-end learning.

\subsection{Group-Equivariant Convolutional Neural Networks}
\textit{Group-equivariant convolutional neural networks} (G-CNNs), first proposed by \cite{GroupEquivariantCNNs}, enforce equivariance to symmetric transformations by replacing the standard convolution with a generalized form, called group convolution. In this section, we first introduce the group convolution by deriving it from the standard convolution. Afterwards, we introduce papers investigating the theory behind group equivariant neural networks. Finally, we present applications of the group convolutional framework to different transformation groups and input domains.

The discrete standard convolution acting on regular 2D input data is defined as\footnote{Precisely, this formula describes the discrete standard correlation. In neural network literature however, it is constantly referred to as convolution since its slightly easier implementation and the only difference being a flipped kernel. We will follow this nomenclature and subsequently call the presented operations convolution, although they technically are correlations.}
\begin{equation}
(f\star k)(u) = \sum_{v\in\mathbb{Z}^2} f(v)k(v-u) \text{,} 
\label{eq:Conv}
\end{equation} 
where the convolution kernel $k$ of size $v$ is shifted (translated) over the domain $\mathbb{Z}^2$ of the input function $f$. Usually, the kernel size is significantly smaller than the input size, which leads to two advantages: First, the number of parameters learned during training is reduced compared to a fully connected layer. Additionally, the same kernel is applied at each input position, which is called spatial parameter sharing and induces equivariance to translations $L_t$
\begin{align}
	\begin{split}
		[[L_t f]\star k](u) &= \sum_{v}f(v-t)k(v-u) \\
		&= \sum_{v}f(v)k(v-(u-t)) \\
		&= [[L_t[f\star k]](u)  \text{.}
	\end{split}
	\label{eq:Proof_Equivariance}
\end{align}

\textbf{Group convolutions} are a generalization of the standard convolution for arbitrary transformation groups G. The continuous group convolution is defined as
\begin{equation}
\label{eq:GConvCont}
(f \star_G k)(u) = \int_{G}f(g)k(u^{-1}g)\; d\mu(g)\text{,}
\end{equation}
where $\mu$ is the Haar measure.
The shift, i.e. the action of the translation group, in Formula \ref{eq:Conv} is replaced with the action $g$ of the transformation group $G$. The discrete group convolution is defined as

\begin{equation}
(f\star_G k)(u) = \sum_{g\in G}f(g)k(u^{-1}g) \text{.} \label{eq:GConv}
\end{equation}
It is easy to see, that the group convolution acts on the domain of the group $G$. Consequently, its input and filters are defined on $G$ as well. 
The group convolution defined for $G$ is equivariant w.r.t. transformations $L_h$ of that group
\begin{align}
	\begin{split}
		[[L_h f]\star_G k](u) &= \sum_{g\in G}f(h^{-1}g)k(u^{-1}g) \\
		&= \sum_{g\in G}f(g)k((h^{-1}u)^{-1}g) \\
		&= [[L_h[f\star_G k]](u)  \text{.}
	\end{split}
\end{align}

Since the input data is usually not defined on the group domain (e.g. images lie on a regular grid), a \textbf{lifting layer} needs to be defined, which lifts the input, e.g. from $\mathbb{Z}^2$ in images, to the group domain
\begin{equation}
(f\star_G k)(u) = \sum_{v \in\mathbb{Z}^2} f(v)k(u^{-1}v) \text{,} \label{eq:GConvLifting}
\end{equation}
where $(f\star_G k) \in G$ and $k$ is a learnable filter defined in the input domain $\mathbb{Z}^2$. 

\subsubsection{Theoretical Aspects}
In general, CNNs do not only consist of multiple linear filter operations, but also use non-linearities. If each operation enforces or preserves equivariance, the entire DNN is equivariant. \cite{GroupEquivariantCNNs} prove that point-wise non-linearities and pooling operations acting on a region which is a subgroup $H\subset G$ preserves equivariance of the network. Hence, it is sufficient to prove equivariance for the convolutional layers used within a G-CNN to prove equivariance for the entire G-CNN, if no other non-linearities are used. 

While \cite{GroupEquivariantCNNs} primarily propose to use G-CNNs focused on an application, i.e. for rotation equivariant classification (see Section \ref{sec:GCNNApps}), several works investigate general theoretical frameworks that enable the classification of existing G-CNNs as well as the expansion towards more complex groups or other input domains, e.g. graphs \citep{SteerableCNNs, Kondor, GeneralEquivariantCNNs, EstevesSurvey}. 

\cite{Kondor} provide a mathematical prove that a DNN with linear maps is equivariant to the action of a compact group $G$ of its input if and only if each layer implements the generalized group convolution defined in Equation \ref{eq:GConv}. For their proof, they use concepts from group representation theory. Additionally, it is proven that those convolutions can be defined over quotient spaces of the transformation group $G$ because the network activations are usually defined on homogeneous spaces associated to the group. Finally, the benefits of calculating convolutions using irreducible Fourier space representations are emphasized which lead to sparse matrix multiplications when defined on the appropriate quotient spaces. 

\cite{GeneralEquivariantCNNs} expand Kondor and Trivedi's theory by defining convolutional feature spaces as fields over a homogeneous \textit{base space}. This enables a systematic classification of various equivariant convolutional layers based on their symmetry group, the base space the features are attached to and the field type, which can be regular, trivial or irreducible. The field type is closely related to group representation theory and describes, how the values change under group transformations of the input. Trivial representations are invariant, i.e. the individual values do not change under transformations, the values of regular fields are shifted along the group dimension and irreducible fields change according to the irreducible representation of the transformation group. Finally, \cite{GeneralEquivariantCNNs} use the Mackey theory to prove that a convolution with equivariant kernels is not only a necessary and sufficient condition for equivariance, but that this is also the most general class of equivariant linear maps. Finally, they propose a general framework on how to parameterize the learnable equivariant convolutional kernels within DNNs.

\cite{EstevesSurvey} gives a broad overview over the mathematical concepts underlying group equivariant CNNs, which are: group representation theory, integration and harmonic analysis on non-Euclidean spaces, and differential geometry. Esteves shows applications of his theory for the three-dimensional group of rotations $\mathbf{SO}(3)$ and elucidates the relation between spherical harmonics and the irreducible representations which arise when calculating the convolution in the Fourier domain. Finally, based on results from group representation theory, both Kondor and Trivedi's and Cohen et al.'s definition of group equivariant CNNs are discussed.

\cite{EquivarianceParameterSharing} prove that a DNN is equivariant w.r.t. a group $G$, iff the group action explains the symmetries within the network's parameter matrices. Based on this insight, two parameter-sharing schemes are provided to enforce the desired in- or equivariance properties. Finally, it is shown that the weight-sharing approach guarantees sensitivity to other permutation groups.

\cite{SteerableCNNs} propose a general framework where the features of an equivariant CNN are defined as steerable compositions of feature types each encoding certain transformation symmetries. This implies that the learned representations transform in a predictable linear manner under transformations of the input. The choice of elementary feature types imposes constraints on the network weights and architecture which reduces the number of learnable parameters. Concepts from group representation theory are used to describe the learned maps of equivariant CNNs. Feature maps are modeled as fibers at each position of the base space. An equivariant map processing those fibers is called \textit{intertwiner} and can be expressed as a direct sum of \textit{irreducible representations}. Finally, they elaborate that the steerable CNN framework can be expanded to other forms of homogeneous spaces, e.g. graphs. 

\cite{PACBayesianGeneralizationBoundEquiv} derive a PAC-Bayesian generalization bound for steerable CNNs whose representations can be expressed as a direct sum of irreducible representation. The bound combines the representation theoretic framework from \cite{SteerableCNNs} with the PAC-Bayes framework in the Fourier domain and can be used to determine the impact of group size, multiplicity and type of irreducible representations on the generalization error. In general, the generalization capability of G-CNNs improves for larger group sizes. 

\cite{ClassificationGInvariant} classify all single-hidden-layer DNNs with ReLU-activations that are invariant to a finite orthogonal group $G$. Therefore, the \textit{signed permutations} of the architectures' underlying irreducible representations is used. This systematic G-invariant architecture classification allows to find the optimal invariant architecture for a specific problem. In combination with a characterization of network morphisms, i.e. transitions between different architectures, Neural Architecture Search among different G-invariant architectures is enabled.

\subsubsection{Applications}\label{sec:GCNNApps}
In this section, we present applications of G-CNNs ordered by their equivariance group.  Additionally, we categorize the approaches following the procedure proposed by \cite{GeneralEquivariantCNNs} (see Table \ref{tab:Equi}).

\paragraph{2D rotations, translations \& flips} 
First, we present approaches equivariant to 2D translations, rotations and flips. Rotations and translations alone form the group $\mathbf{SE}(2)$, while all three transformations correspond to $\mathbf{E}(2)$. \cite{GroupEquivariantCNNs} apply the group convolution to the discrete wallpaper subgroups $p4$ and $p4m$ consisting of 2D translations, $90^\circ$ rotations and flips.The learned filters are transformed with all possible rotations and flips and stacked along the \textit{group dimension}, which corresponds to the regular representation. The standard convolution can then be applied to obtain the desired equivariance.

\cite{Hoogeboom} expand this framework to $60^\circ$-rotations by re-sampling the image on a hexagonal grid via bi-linear interpolation instead of using square pixels. The convolutional filters and outputs are directly defined on the hexagonal grid which allows a equivariant network without additional interpolation. 

\cite{Bekkers} further generalize this approach to any discrete group $\mathbf{SE}(2, N) \leqslant \mathbf{SE}(2)$, which includes $N$ two-dimensional rotations. Bi-linear interpolation is applied to rotate the filters beyond 90$^\circ$ rotations that are fully covered by the regular grid. In their experiments, the best results are achieved using $N=8$ or $N=16$ rotation angles.

Rotation-equivariant Vector Field Networks \citep{MarcosRot} reduce the parameter and memory consumption of rotation-equivariant CNNs by only storing the maximum activation and the orientation that caused it at each location. This can be seen as a vector field where the activation is responsible for the vector's length while the orientation is obtained from the corresponding rotation of the filter. The authors conclude that only storing dominant responses and orientations is sufficient to solve in- and equivariant tasks. Storing an amplitude and orientation is closely related to the irreducible representation of 2D rotations.

\cite{HarmonicNetworks} train a CNN where the filters are restricted to circular harmonics such that the convolution's output is equivariant w.r.t. continuous rotations. In particular, the filters are defined in the polar domain
\begin{equation}
\mathbf{W}_m(r,\phi,R,\beta) = R(r)e^{i(m\phi+\beta)} \text{,}
\end{equation}
where $r$ and $\phi$ are the radius and angle of the polar representation, $m$ is the rotation-order while $R$ and $\beta$ are the learnable radial profile and phase offset. The filters are defined in the complex domain which automatically induces a complex output of the convolution operation that can easily be calculated using four standard convolutions. While the output's modulus is invariant to rotations of the input, the phase stores the equivariant information about its orientation. Harmonic networks are a group-convolution operating on the irreducible representation of the two-dimensional rotation group, which is characterized by the circular harmonics. 

A variety of other approaches rely on using steerable filters for the convolutional kernels.
As mentioned in section \ref{sec:SteerableFilters}, \textbf{steerable filters} can be calculated in arbitrary rotations using a linear combination of basis filters. Similar to the convolution, this principle can be expanded to arbitrary transformations. Steerable filters provide an attractive alternative to compute arbitrarily transformed filters in equivariant CNNs since they do not suffer from interpolation artifacts. 

\cite{SFCNN} follow this approach for 2D rotations. The convolutional kernels are restricted to a linear combination of steerable filters with learnable linear coefficients and the regular $\mathbf{SE}(2,N)$ group convolution is applied to obtain the desired equivariance.
Additionally, a generalized weight initialization scheme is proposed that further improves the performance of the layer.

\cite{E2STCNNs} generalize steerable CNNs to the Euclidean group $\mathbf{E}(2)$. The equivariant convolution kernels of $\mathbf{E}(2)$-CNNs are constrained through the symmetries they are designed to express. Those constraints can be formulated using the \textit{irreducible representations} of the symmetry groups. Hence, a basis of equivariant kernels can be calculated by solving the irrep constraints, reverting the change of basis and taking the union over the different blocks. This results in a general formulation of the kernel constraint using a Fourier basis which leads to harmonic basis elements. Additionally, the proposed framework allows to relax group restrictions for higher layers of G-CNNs which might not benefit from full symmetry.

In contrary to the steerable filter approach, \cite{Diaconu1} propose to replace the interpolation needed to obtain filters at arbitrary rotations by learning a filter basis and all its rotated versions through rotation-invariant coefficients. Therefore, they propose the \textit{unitary group convolution}:
\begin{equation}
[f\star_G k](g) = \sum_{x\in X} f(x)L_g[k](x) \text{,}
\end{equation}
which is only equivariant, if the inner product is unitary, i.e. it is constant if both $f$ and $k$ are transformed by the same transformation group element $g$. The rotated filters are then learned using a novel equivariance loss. The unitary group convolution shows improved robustness to rotated activations and guarantees a better representation stability than the group convolution. 

\cite{ECCO} introduce a rotation-equivariant continuous convolution used for trajectory prediction in order to enable physically consistent predictions. Therefore, a novel weight-sharing scheme based on polar coordinates with orbit decomposition is introduced that obtains equivariance via torus-valued kernels.  

\cite{EquivariantStochasticFields} incorporate equivariance into Gaussian and Conditional Neural Processes. Therefore, they develop a kernel constraint for the Gaussian processes while using $\mathbf{E}(2)$-equivariant steerable filters \citep{E2STCNNs} for the learnable decoder of the Conditional Neural Process. Additionally, they prove that equivariance in the posterior is equal to invariance in the prior data distribution.

\cite{EquivariantGANs} use regular group-convolutions equivariant to flips and 90$^\circ$-rotations in Generative Adversarial Networks to obtain an equivariant latent representation while training the network with an improved sample efficiency. 

\cite{TranslationRotVAE} leverage equivariance to rotations and translations to disentangle the latent representations learned by Variational Autoencoders (VAEs). The encoder network learns a translation-equivariant component, a rotation-equivariant orientation component as well as a translation- and rotation-invariant object representation. A spatially equivariant generator network can then be used to achieve a fully equivariant VAE that is able to learn accurate object representations even from heavily transformed input images.

In a similar fashion, \cite{WinterInvariantAE} propose a general framework to learn group in- and equivariant representations in an unsupervised manner with a encoder-decoder network. The latent representation is again separated into an invariant part learned by a $G$-invariant encoder-decoder pair, and the corresponding equivariant group action needed to recover the input in the correct orientation learned by a \textit{suitable group function}. Moreover, general conditions for any group $G$ and experiments for rotations, translations and permutations using an Autoencoder are provided.

\begin{table*}[t]
	\centering
	\resizebox{\columnwidth}{!}{
	\begin{tabular}{cccc}
		\toprule
		Reference & Equivariance group & Base space & Vector field type \\
		\midrule
		\cite{Fukushima, LeCun} & $\mathbb{Z}^2$ & $\mathbb{Z}^2$ & regular \\
		\cite{GroupEquivariantCNNs} & $p4, p4m$ & $\mathbb{Z}^2$ &  regular \\
		\cite{SteerableCNNs} & $p4, p4m$ & $\mathbb{Z}^2$ &  irrep \& regular \\
		\cite{Diaconu2} & $p4$ & $\mathbb{Z}^2$ & regular \\
		\cite{RomeroCoAttentive} & $p4$ & $\mathbb{Z}^2$ & regular \\
		\cite{Bekkers} & $\mathbf{SE}(2, N)$ &  $\mathbb{Z}^2$ & regular \\
		\cite{MarcosRot} & $\mathbf{SE}(2, N)$ &  $\mathbb{R}^2$ & irrep \& regular \\
		\cite{SFCNN} & $\mathbf{SE}(2, N)$ &  $\mathbb{R}^2$ & regular \\
		\cite{Diaconu1} & $\mathbf{SE}(2, N)$ & $\mathbb{R}^2$ & regular \\	 
		\cite{HarmonicNetworks} & $\mathbf{SE}(2)$ & $\mathbb{R}^2$ & irrep \\
		\cite{ECCO} & $\mathbf{SE}(2)$ & $\mathbb{R}^2$ &  trivial \& regular \\
		\cite{E2STCNNs} & $\mathbf{E}(2)$ & $\mathbb{R}^2$ & any \\
		\cite{SphericalCNNs} & $\mathbf{SO}(3) $ & $S^2$ & regular \\
		\cite{Esteves, SphericalHourglass, EquivariantMultiView} & $\mathbf{SO}(3) $ & $S^2$ & trivial \\
		\cite{Perraudin} & $\mathbf{SO}(3) $ & $S^2$ & trivial \\
		\cite{DefferrardDeepSphere} & $\mathbf{SO}(3) $ & $S^2$ & trivial \\
		\cite{ClebschGordanNets} & $\mathbf{SO}(3)$ & $S^2$ & irrep \\  
		\cite{Jiang} & $\mathbf{SO}(3) $ & $S^2$ & irrep \\
		\cite{SpinWeightedSpherical}  & $\mathbf{SO}(3)$ & $S^2$ & irrep \\ 
		\cite{Winkels} & $D_4$, $D_{4h}$, $O$, $O_h$ & $\mathbb{Z}^3$ & regular \\
		\cite{CubeNet} & $S_4$, $T_{4}$, $V$ & $\mathbb{Z}^3$ & regular \\
		\cite{3DSTCNNs} & $\mathbf{SE}(3, N)$ & $\mathbb{R}^3$ & irrep \\ 
		\cite{LocalRotationInvariance3D}  & $\mathbf{SE}(3, N) $ & $\mathbb{R}^3$ & regular \& trivial \\
		\cite{XuScale} & $\mathbb{R}^2 \rtimes S$ & $\mathbb{R}^2 $ & regular \\
		\cite{LocalScaleInvariance} & $\mathbb{R}^2 \rtimes S$ & $\mathbb{R}^2 $ & trivial \\ 
		\cite{MarcosScale} & $\mathbb{R}^2 \rtimes S$ & $\mathbb{R}^2 $ & irrep \& regular \\ 
		\cite{Worrall19} & $\mathbb{R}^2 \rtimes S$ & $\mathbb{R}^2 $ & regular \\ 
		\cite{ZhuScale} & $\mathbb{R}^2 \rtimes S$ & $\mathbb{R}^2 $ & regular \\ 
		\cite{GhoshScale} & $\mathbb{R}^2 \rtimes S$ & $\mathbb{R}^2 $ & regular \& trivial \\ 
		\cite{Sosnovik, DISCO} & $\mathbb{R}^2 \rtimes S$ & $\mathbb{R}^2 $ & regular \\
		\cite{IssakkimuthuS19} & $\mathbf{SE}(2) \rtimes S$ & $\mathbb{R}^2 $ & irrep \\ 
		\cite{GIFT} & $\mathbf{SE}(2) \rtimes S$ & $\mathbb{R}^2$ & regular \& trivial\\
		\cite{TensorFieldNetworks} & $\mathbf{SE}(3) $ &   $\mathcal{G}$ & irrep \& regular \\
		\cite{Batzner}  & $\mathbf{E}(3)$&   $\mathcal{G}$ & irrep \\ 
		\cite{E3MessagePassing}  & $\mathbf{E}(3)$ &  $\mathcal{G}$  & irrep \\ 
		\cite{ENEquivariantGNNs}  & $\mathbf{E}(N)$ &   $\mathcal{G}$ & trivial \\
		\cite{Horie}  & $\mathbf{E}(3)$ &  $\mathcal{G}$ & trivial \\ 
		\cite{Gauge} & Local Gauge $G$ & Manifold & irrep \\ 
		\cite{GaugeEquivariantMeshCNNs} & Local Gauge-$G$ & Mesh & irrep \\ 
		\cite{Shakerinava} & $\mathbf{SE}(3) \rtimes S$ \& Local Gauge-$G$ & $S^2$ & irrep \\
		\cite{BekkersLie} & $\mathbb{R}^2 \rtimes H$ & $\mathbb{R}^2 $ & regular \\ 
		\cite{Finzi2020} & $G$ & $\mathbb{R}^n $ & regular \\ 
		\cite{Finzi2021} &$ \mathbf{GL}(n)$ & $\mathbb{R}^n $ & any \\ 
		\cite{WignerEckartGConv} & G & X & any\\
		\cite{EnSTCNNs} & G & X & any \\
		\bottomrule
	\end{tabular}
	}
	\caption{Overview of in- and equivariant CNNs classified by the equivariance group $G$, the base space $G/H$ and the field type $\rho$. Based on \cite{GeneralEquivariantCNNs}.}
	\label{tab:Equi}
\end{table*}

\paragraph{3D rotations and translations}
A variety of approaches expand equivariance towards 3D signals and groups. Since finite groups grow exponentially with increasing dimension, sophisticated analytical solutions for equivariance are even more crucial than in the 2D case. Especially for larger groups, irreducible representations provide a more efficient way to represent the group domain outputs than the regular representation which grows with the group size. 

First, we present work which achieves equivariance to the group of 3D rotations SO(3) for inputs defined on the sphere $S^2$, e.g. spherical images or global climate models.
\cite{SphericalCNNs} propose a spherical correlation incorporating equivariance to SO(3). Therefore, the signal is transformed with the generalized Group Fast Fourier transformation. The convolution can then be efficiently computed in the Fourier domain using a multiplication. Finally, the inverse Group Fourier transformation is applied to obtain the final result.

\cite{Esteves} independently suggest to calculate the spherical convolution in the Fourier domain but use trivial instead of regular group representations. Both apply the spherical convolution to 3D shape recognition. \cite{SphericalHourglass} use the spherical convolutional layer to enforce invariance to the camera-pose in a DNN for semantic segmentation of spherical images. \cite{EquivariantMultiView} create a group representation from multiple views of a single object and process it with spherical CNNs in order to encode shape information equivariant to the icosahedral group. Additionally, they use the log-polar transformation, where in-plane rotations act as translations, which allows to generalize from less viewpoints by exploiting their equivariance properties.

\cite{Spezialetti} use spherical convolutions to learn SO(3)-equivariant feature descriptors for 3D shape correspondence in an unsupervised manner. The learned robust representation is combined with a orientation estimation via an external local reference frame at test time to learn effective 3D shape descriptors. 

\cite{Perraudin} approximate the spherical convolution by treating the discrete sphere as a graph and applying a graph convolution. Their approach, called DeepSphere, achieves equivariance to 3D rotations by restricting the learned filters to be radial and is applied to cosmology images. While this approach fails to achieve exact equivariance, it allows to significantly reduce the computational complexity of spherical convolutions. In a subsequent paper, \cite{DefferrardDeepSphere} propose minor improvements of DeepSphere and show that the number of connected neighbors on the spherical graph can be used to trade-off between equivariance guarantees and computational complexity.

Comparably, \cite{Jiang} process the sphere as a unstructured grid (mesh) and apply a mesh convolution. Rotation-equivariance is guaranteed by using a linear combination of parameterized differential operators. While the weights of the linear combination are learnable, the differential operators are efficiently estimated using one-ring neighbors. The operation itself is applicable to arbitrary unstructured grids and shows promising results for tasks in the spherical domain while being parameter-efficient.

\cite{ClebschGordanNets} further generalize spherical SO(3)-equivariant CNNs by proposing to use the Clebsch-Gordan transformation as a general purpose nonlinearity for rotation-equivariant spherical CNNs. The Clebsch-Gordan transformation decomposes a tensor into a product of its irreducible representations. It was first used in neural networks for rotation-symmetric physical systems by \cite{Kondor1}. Since the transformation is calculated directly in the Fourier space, it avoids the frequent calculation of forward and backward Fourier transformations after each layer. Moreover, \cite{ClebschGordanNets} generalize the Fourier transformation to compact continuous groups.

\cite{SpinWeightedSpherical} introduce spin-weighted spherical CNNs which are SO(3)-equivariant by using complex-valued spin-weighted spherical functions for the learnable filters. Thereby, the computationally expensive lifting of the input to SO(3) is avoided. On the other hand, the resulting filters are more expressive than their scalar isotropic counterparts that achieve equivariance directly operating on the sphere. Similar to the Harmonic Networks for 2D rotations \citep{HarmonicNetworks}, the responses are complex-valued vector fields where an input rotation induces a phase shift. The responses are directly computed in the spectral domain and achieve continuous SO(3)-equivariance.

\cite{RotationEquivariantNaturalIllumination} expand vector neurons towards rotation-equivariant conditional neural fields for spherical images. Equivariance to rotations around the vertical axis is achieved via a conditional latent code that represents the desired direction. Combined with a variational auto-decoder and statistical priors about natural lighting conditions their method is successfully applied to inverse rendering tasks.

Multiple proposals achieve equivariance to the group of 3D rotations and translations SE(3) for three-dimensional inputs. \cite{Winkels} transfer the group convolution of \cite{GroupEquivariantCNNs} to 3D by transforming the filters using the finite 3D groups $D_4, D_{4h}, O$ and $O_h$ which describe the symmetries of cuboids and cubes, respectively. \cite{CubeNet} independently propose the same procedure using the cube group $S_4$ containing 24 right-angle rotations, the tetrahedral group $T_4$ including the 12 even rotations and Klein's four-group $V$, which is commutative and the smallest non-cyclic group.

\cite{3DSTCNNs} expand the theory of steerable equivariant CNNs to 3D. They propose to parameterize 3D rotation steerable convolutional kernels as linear combinations of spherical harmonics with learnable weights. Additionally, they show that only the angular part of the spherical harmonics is restricted by the equivariance constraints, while the radial part can be chosen arbitrarily. 

\cite{LocalRotationInvariance3D} introduce three different networks that achieve local 3D rotation invariance and global rotation equivariance at the same time. Therefore, they use a convolution based on rotated filters, rotation-steerable filters and learned solid spherical energy invariants. Specifically, they only use a single lifting layer and orientation pooling followed by global average pooling to obtain local invariants defined by the kernel size that are further processed by fully connected layers. Using the same architecture, local invariants outperform their global counterparts for a medical image analysis task.

\cite{E3NN} provide a general pytorch-based framework to compute any E(3)-equivariant CNN.

\paragraph{Scale} Another interesting transformation that naturally occurs in images is scale, which for example results from variable camera-to-object distances. Consequently, enforcing group-equivariance to scaling transformations is beneficial for Computer Vision tasks. Compared to rotations, enforcing equivariance to scales is more challenging since the scale group is non-cyclic and unbounded. Additionally, scaling can only be modeled as a semi-group due to the information loss when down-scaling an image which makes the group action non-invertible. In mathematical terms, scale transformations are often called dilation.

\cite{XuScale} use a similar approach to G-CNNs to obtain scale invariance in a convolutional neural network which they call multi-column CNN. The core idea is to share scaled versions of the same convolutional filter among different columns, each resulting in a maximum response at a different scale of the same pattern. The scaled versions of the filters are calculated using bi-linear interpolation for upscaling while the minimum $L_2$-norm filter is used for downscaling. The columns, which can be seen as the group transformation channel in the G-CNN framework, process the input independently, resulting in a column-flip when a pattern is scaled. Finally, the column activations are concatenated and processed by classification layers.

Conversely, \cite{LocalScaleInvariance} process multiple scaled versions of the same input obtained using bi-linear interpolation at each convolutional layer. At each location, the maximum response is kept by applying the max pooling operation over all scales. 
Thereby, each layer guarantees local scale-invariance. While this approach leads to promising results, it involves two transformations of the input per convolution layer. \cite{MarcosScale} expand this approach by additionally storing the information about which scale caused the maximum response at each location in a vector field. Thereby, they disentangle the scale and the magnitude of the responses. Higher convolutional layers process both information at once using a vector field convolution. 

\cite{Worrall19} propose to formally enhance the group convolution such that it can be applied to transformations modeled as a semigroup
\begin{equation}
[k \star_S f](s) = \sum_{x\in X} k(x)L_s[f](x) \text{.}
\end{equation}
In comparison to the group convolution (Equation \ref{eq:GConv}), the signal is transformed instead of the filter, which is similar to the approach used by \cite{LocalScaleInvariance}. Again, a transformation acting on the input induces a shift in the semi-group convolution's output. By defining the input images as scale-spaces, i.e. the input and multiple blurred versions of it, and modeling dilations as a discrete semigroup, a scale equivariant NN can be defined. \cite{Worrall19} restrict their approach to integer scaling to avoid interpolation when creating the scale space, which on the downside leads to unwanted boundary effects for non-integer scalings of the input. 

\cite{ZhuScale} propose a scale-equivariant CNN using joint convolutions across the space and scaling group. In order to reduce the model complexity, they decompose the convolutional filters under two pre-fixed separable Fourier-Bessel bases with trainable expansion coefficients. At the same time, they truncate the filters to low frequency components which leads to an improved deformation robustness and a reduced parameter consumption.

\cite{GhoshScale} adapt the local scale-invariant convolutions of Kanazawa et al. by transforming scale-steerable filters instead of the signal itself. The filters are composed of linear combinations of log-radial harmonics, an adapted version of the circular harmonics. By using the steerable filter approach, interpolation artifacts are avoided which leads to an improved scale robustness and performance. Additionally, the second scale operation of Kanazawa's approach is avoided by transforming the filters instead of the in- and output.

\cite{Sosnovik} use steerable filters and scale-equivariant group convolutions in combination with a discrete number of scales. This approach allows to efficiently calculate filters for arbitrary real-valued scaling factors. Two-dimensional Hermite polynomials with a Gaussian envelope are used as the steerable filter basis. As usual for steerable filter CNNs, the filter basis is pre-computed while the weights of the linear combinations are learned. While the network could in theory learn from inter-scale interactions, the experiments show that the DNNs perform better using limited inter-scale interactions. This principle was applied to scale-equivariant siamese trackers for object localization in \cite{ScaleTracking}. Scale-equivariance helps to improve the performance of object trackers through a better scale estimation as well as a better notion of similarity between objects of different scales. \cite{DISCO} replace the fixed basis with an approximation whenever no closed-form solution can be computed, i.e. for non-integer scales or due to mapping continuous kernels to a discrete grid. The basis is optimized by directly minimizing the equivariance error, which further improves the performance of scale-equivariant steerable convolutional networks.

\cite{WaveletNetwork} introduce scale-and-translation equivariant networks applied to time-series which naturally results in wavelet filters. This approach achieves performance comparable to hand-designed spectral methods when applied directly on raw time-series data.

\cite{IssakkimuthuS19} propose to learn rotation-equivariant filter bases using an autoencoder with $\mathbf{SO}(2)$-equivariant mapping and tensor nonlinearities. Additionally, scale-coupled bases are used in combination with the tensor nonlinearity to obtain scale-robust filter bases. Finally, they show that the learned bases closely resemble their Fourier counterparts and allow for good performance in classification tasks.

\cite{GIFT} use a two-step approach to obtain pixel-wise visual descriptors invariant to scale and rotations, which can be used to find correspondence points in a set of images. First, they process transformed versions of the input image using a standard CNN. Second, they use the group convolution for rotations and scales as well as bilinear group pooling to obtain invariant features, which they call GIFT descriptors.

\paragraph{Graph Neural Networks}
Graph Neural Networks (GNNs) are deep neural networks designed to operate directly on data structured as graphs $\mathcal{G}$. By design, GNNs are permutation-in- or equivariant, i.e. the order of graph nodes does not affect the desired output. The graph convolution is usually computed in the spectral domain \citep{GraphNNs} with localized filters \citep{DefferardGraphNNs} and can be approximated in an compute-efficient way \citep{KipfGraphNNs}. In a more general form, GNNs with multiple stacked layers can also be interpreted as a message passing DNN, since the information is aggregated from the neighboring nodes at each layer. Message passing GNNs can be convolutional, but can also use other functions for feature aggregation. In- or equivariance to other transformations than permutations generally improves the sample complexity and performance of Graph Neural Networks.

\cite{TensorFieldNetworks} use a mix of continuous convolutions and filters restricted to spherical harmonics to learn representations that are locally equivariant to 3D rotations, translations and point permutations on point clouds, i.e. unconnected graphs. Point-wise convolutions are used to process vector fields defined on each point. The convolution is calculated using the equivariant tensor product via the Clebsch-Gordan coefficients and the irreducible representations of $\mathbf{SO}(3)$.

\cite{Batzner} enable data-efficient learning of interatomic potentials with E(3)-equivariant Graph convolutions. In contrast to other GNNs for molecular dynamic simulations, a relative distance vector and tensor-valued node features are used instead of scalar distances and features to encode positional information. E(3) equivariance is achieved by restricting the convolutional filters to spherical harmonics with learnable radial profile and phase offset. 

\cite{E3MessagePassing} introduce steerable GNNs that compute node and edge attributes that achieve equivariance to E(3) rather than invariance. Their non-linear generalization of E(3)-steerable group convolutions computed via the Clebsch-Gordan tensor product can be incorporated into both the message passing and the update function of GNNs. A general class of equivariant activation functions for steerable feature fields is proposed. Finally, the benefits of non-linear message aggregation compared to linear point convolutions as well as of equivariant, steerable message passing compared to invariant messages is demonstrated.

\cite{ENEquivariantGNNs} adapt the graph convolution layer such that it achieves E(n) equivariance without using tensor-valued intermediate representations or spherical harmonics. Equivariance is achieved with an updated edge operation that considers the relative squared distance between nodes and sequentially updating the relative position of each particle with the weighted sum of all radial distances. This approach is easier to compute and enables scaling beyond 3 dimensions. 

\cite{Horie} compute graph features in- and equivariant to isometric transformations in a computationally efficient manner by tweaking the adjacency matrix of Graph Neural Networks. Their isometric graph CNNs achieve good performance on geometrical and physical simulation data. Additionally, the proposed networks can be used to replace conventional physical models such as Finite Element Analysis since inference is significantly faster than related equivariant graph NNs which allows to scale-up to the large graphs needed for physical models.

\paragraph{Arbitrary Groups \& Inputs}
\cite{Gauge} extend group convolutions to local gauge transformations, which enables equivariant CNNs on manifolds instead of on homogeneous base spaces. The convolutions only depend on the intrinsic geometry of the manifold. Compared to previous work, equivariance is guaranteed w.r.t. local transformations instead of global ones. This framework is applied to signals on a icosahedron which is an approximation of the sphere.

\cite{GaugeEquivariantMeshCNNs} further extend this approach to meshes. Whereas most related work simply treats meshes as a graph, the features of gauge equivariant mesh CNNs are able to capture the mesh geometry, e.g. the orientations of neighbor vertices. Therefore, graph convolutions using anisotropic kernels are defined that generate gauge-equivariant features passed within the CNN using parallel transport. 

\cite{Shakerinava} implement a hierarchy of symmetries that involves both local gauge transformations and global rotations and scalings of the pixelized spheres also covering interactions between them. Therefore, they propose equivariant maps based on systems of blocks and a novel equivariant padding.

\cite{Finzi2020} propose a convolutional layer equivariant to transformations which can be expressed as a Lie group based on a surjective exponential map. The novel layer can be applied to arbitrary continuous data, including regular grids (images) and point clouds. Equivariance to a new group can be achieved by implementing the group exponential and logarithm maps and using the general layer framework. Hence, the proposed layer is flexible and can be used for a variety of different problem settings.

Comparably, \cite{BekkersLie} introduces a convolutional layer equivariant to groups that are a semidirect product of the translation group and arbitrary Lie groups. Therefore, the convolutional kernels are parameterized using B-spline basis functions defined on the Lie algebra of the corresponding Lie group. By altering the respective type of the B-spline basis functions localized, dilated and deformable group convolutions can be implemented. 

\cite{WignerEckartGConv} provide a general solution to find G-steerable kernels for arbitrary compact groups $G$ by generalizing the Wigner-Eckart theorem. CNNs equivariant to translations and $G$ can be built by applying convolutions with G-steerable kernels. The equivariant kernels consist of learnable endomorphism bases, Clebsch-Gordan coefficients and harmonic basis functions characterized by the group G and the corresponding homogeneous space $X$.

\cite{EnSTCNNs} expand the previous approach to more general, not necessarily homogeneous spaces. Therefore, the harmonic basis functions defined on the orbits of $G$ are replaced with a G-steerable basis $\mathcal{B}$ defined over the whole space. This allows to calculate unconstrained scalar filters with controllable bandwidth and aliasing properties. Additionally, the discretization is disentangled from the choice of G, which allows for an easier implementation for new groups. 

\cite{EquivariantConditionalNeuralProcesses} propose a group-equivariant conditional Neural Process used for meta-learning that encodes transformation equivariance and permutation invariance at the same time. While the encoder induces a functional equivariant representation, Lie-Convs \citep{Finzi2020} are used in the decoder to achieve equivariance in the output space.

\cite{LieGroupVAE} use an exponential map to allow group representations instead of standard vector spaces for the latent space of VAEs. Thereby, equivariance is enforced for unsupervised disentanglement learning. A Lie algebra parametrization is used to convert the training problem to linear spaces which allows to apply general optimization methods. Furthermore, commutative decomposition constraints encouraging disentanglement are derived. 

\cite{Finzi2021} provide a general algorithm to design equivariant multi-layer perceptrons (MLPs) for any arbitrary representation and matrix group. Therefore, they solve the kernel equivariance constraints via a singular value decomposition of infinitesimal or discrete Lie algebra generators. While this approach provides a more general solution than convolutions, its computation is slower due to the required dense matrix multiplications.

In summary, group convolutions have been applied to a variety of transformations in both two and three dimensions. While most applications thus far are limited to simple transformations on homogeneous base spaces, e.g. scale and rotations, current research investigates how to generalize the group convolution to more complex groups and input domains. 

\subsection{Non-Linear Equivariant Maps}
While group-equivariant convolutional neural networks are the most general linear map that guarantee equivariance, non-linear operations such as self-attention used for transformers have recently gained significant research interest \citep{Transformer, VisualTransformer}. One interesting field is thus, how those non-linear operations can be adapted such that they guarantee in- or equivariance. Additionally, the general non-linearities which are also used for CNNs have been investigated.

\cite{Diaconu2} investigate combining the roto-translation group convolution with a self-attention mechanism to obtain equivariant data-dependent filters. The proposed networks achieve improved results compared to self-attention or equivariant convolutions on their own while reducing the number of model parameters - allowing to use data-dependent filters and incorporating geometrical prior knowledge at once.

Inspired by the human visual system which is not fully rotation equivariant, \cite{RomeroCoAttentive} propose to use an attention mechanism to learn co-occurring transformations within the dataset. Thereby, the network can exploit information about co-occurring transformations without disrupting its global equivariance. For example, a network for face detection should exploit the relative orientation between the eyes and the nose. This concept improves discrete rotation equivariant CNNs for both full and partial rotation equivariant tasks.

\cite{GroupSelfAtt} provide a general group-equivariant formulation of the self-attention layer. Equivariance is achieved by modifying the relative positional encodings such that they are invariant to the group. Additionally, the computed representations are lifted to the group such that they can express equi- and not only invariance, similar to regular group convolutions. While group-equivariant self-attention based networks achieve better performance than their non-equivariant counterparts, they fail to outperform their convolutional counterparts for small-scale datasets.

\cite{LieTransformer} extend the group-equivariant self-attention layer to general Lie groups as well as general domains using a lifting based approach. They achieve exact equivariance for finite subgroups and equivariance in expectation for general Lie groups by using a Monte Carlo estimate of the involved integral.

\cite{SE3Transformer} introduce a self-attention module equivariant to 3D rotations and translations and apply it to point clouds, i.e. graphs. An adapted Tensor Field Network \citep{TensorFieldNetworks} based on spherical harmonics is used to obtain equivariance. Additionally, each layer learns invariant attention weights for each node combination and a self-interaction term that facilitates to learn node-wise self-attention. An iterative approach allows to use non-fixed basis functions which are required for variable graphs \citep{IterativeSE3Transformer}.

\cite{GaugeEquivariantTransformer} propose a self-attention operator that is equivariant to gauge transformations as well as invariant to global rotations. Gauge equivariance is achieved by constraining the learnable matrices of the attention layer using a Taylor series expansion, while the rotation invariance is obtained by a projection onto local coordinate systems. The parallel transport of feature maps is optimized by using an extension of the regular representation built upon the orthogonal representation obtained via the irreps. Additionally, an equivariance error bound is provided.

\cite{EfficientEquivariantNetworks} propose a general parameter-, data- and compute-efficient equivariant layer that includes both group-self-attention and -convolutions as special cases. The feature aggregation is decoupled into a dynamic kernel generation method based on input features and a feature encoder that provides equivariance by only depending on relative positions of input pairs. 

\subsubsection{Non-Linearities}
In general, the non-linearities of group-equivariant DNNs need to be adapted such that they commute, i.e. do not destroy the equivariance information \citep{GroupEquivariantCNNs}. For the example of regular group representations, the non-linearities need to be applied in a point-wise fashion, which means that the whole group channel needs to be adapted in the same way. Nevertheless, further research investigates the effects of non-linearities and down-sampling on the desired equivariance properties.

\cite{GroupSubsampling} investigate the role of pooling or strided convolutions on the in- and equivariance properties of equivariant neural networks and propose an adapted pooling version that achieves exact in- or equivariance. 

\cite{EquivNonlin} propose non-linearities based on the Fast Fourier Transformation that yield exact equivariance for polynomial non-linearties and approximate solutions with tuneable parameters for other non-linearities. 

To conclude, expansions of the group-equivariant framework towards non-linear maps such as (self-)attention have recently been introduced. This allows to leverage the benefits of guaranteed equivariance for novel, transformer-like architectures. In addition, improved non-linearities that yield exact in- or equivariance rather than sub-sampled approximations have been proposed.

\subsection{Capsules}
Capsules, first introduced by \cite{HintonTransformingAE}, are a specific neural network architecture designed to explicitly learn in- and equivariant representations. A capsule is a group of neurons that perform internal calculations to output a small vector of highly informative outputs. This vector consists of the probability that the visual entity the capsule specializes on is present and its instantiation parameters, e.g. its pose, lightning conditions or deformation. While the former part should be invariant to changes of the instantiation parameters, the latter part should be equivariant. The proposed capsules are trained on pairs of transformed images with known transformation matrix using a transforming autoencoder.
Capsules can be arranged in a \textit{capsule network} similar to conventional DNNs consisting of multiple layers with multiple nodes.
In capsule networks, a \textit{routing algorithm} is used to determine, which lower-layer capsules send information to their higher-layer counterparts.

\cite{DynamicRoutingCapsules} propose a variation of capsules where the instantiation parameters are encoded via the orientation of the capsule's output vector, while its length determines the probability that the entity exists in the input. This enables to use a dynamic routing approach between lower- and higher-level capsules. The routing algorithm is based on calculating the scalar product between the capsules' respective vectors to determine part-whole relationships. Convolutional layers are used within the capsules to benefit from weight sharing. The approach enables capsules to learn the pose information inherently through the routing mechanism. 

\cite{Capsules} introduce a novel type of capsules consisting of a logistic unit indicating the presence of an entity and a pose matrix storing the entity's pose information. The logistic unit allows to optimize an objective function specialized on detecting entities while the pose matrix allows to calculate simplified transformation matrices between capsules. Moreover, a new routing algorithm based on Expectation Maximization (EM) is proposed such that active capsules receive a cluster of similar pose votes from lower-level capsules. 

\cite{CapsuleAutoencoders} train a two-stage capsule autoencoder in a unsupervised manner. The first stage segments an image into object parts and their poses which are then used to reconstruct each image pixel as a mixture of pixels of transformed part templates. The second stage organizes the already discovered parts and poses into a smaller set of objects that are trained on explaining the part poses via a mixture of predictions for each part. When clustering the output of the second stage and assigning the corresponding class labels, state-of-the-art results for unsupervised classification on SVHN is achieved.

\cite{LenssenGroupCapsules} combine capsules with group convolutions to guarantee transformation-invariant output activations and equivariant pose vectors. Therefore, the capsules utilize group convolutions for their calculations, store the pose information as group elements and adapt the dynamic routing algorithm to guarantee equivariance under specified conditions. The group equivariant capsules are applied to SO(2) using spline-based convolutional kernels to avoid interpolation. While the proposed capsules now incorporate guaranteed equivariance properties, restricting the pose vector to group representations prevents the capsule from extracting arbitrary pose information, e.g. lighting.

\cite{VenkDeepEquivariantCapsules} propose space-of-variation capsules in order to improve the scalability of capsule networks and guaranteeing equivariance properties at the same time. Instead of learning pair-wise relationships between capsules, each capsule learns to encode the manifold of legal pose-variations, called space-of-variations, through a neural network that uses group convolutions to benefit from increased parameter sharing. Additionally, a provably equivariant routing procedure guarantees that learned part-whole relationships are preserved under transformations. In comparison to \cite{LenssenGroupCapsules}, the pose information is not embedded with group elements. While this reduces the transformation efficiency of the representation, compositions involving non-geometrical properties can be learned more efficiently. 

To summarize, capsules disentangle visual entities into an invariant presence probability and the corresponding equivariant pose information, which are in their general form learned from data rather than guaranteed. A broad field of research applies capsules to different problem settings or adapts their technicalities, e.g. the routing mechanism. However, listing these proposals is beyond the scope of this survey.

\subsection{Invariant Integration}
The previously presented approaches mainly focus on building equivariant DNNs. Usually, global pooling operations among the group and spatial dimensions are used to obtain the desired invariance properties, e.g. for classification. Nevertheless, alternative approaches obtain guaranteed invariance while adding targeted model capacity to further increase the sample complexity of invariant DNNs.

Invariant Integration is an approach to construct a complete feature space w.r.t. a transformation first applied to pattern recognition tasks by \cite{SM_Existence}. A complete feature space maps all equivalent patterns according to a transformation group $G$ from the signal space $S$ to the same point in the feature space $F$, whereas all non-equivalent patterns are mapped to distinct points. The feature space is invariant but preserves discriminative capacities along other variations. \cite{SM_Existence} mathematically defines conditions for the existence of complete feature spaces in pattern recognition tasks. For finite groups, a complete feature space can be computed by calculating the group average over a function $f$
\begin{equation}\label{eq:groupAverage}
A[f](x) := \int_{G} f(gx)dg \text{.}
\end{equation}
For the choice of the function, \cite{SM_Algos} uses the set of all possible monomials
\begin{equation}
f(x) = x_0^{b_0}x_1^{b_1} \hdots x_{n-1}^{b_n-1} \quad \text{with} \quad b_1 + b_2 + \hdots + b_{n-1} \leq \|G\|
\end{equation}
with monomial exponents $b_i$.
For finite groups, equation \ref{eq:groupAverage} is reduced to a sum over all group elements. The group average using all possible monomials is a valid basis representing the set of invariants. However, it is just an upper bound and computationally inefficient, especially for larger groups. Thus, it is important to reduce the number of monomials by carefully selecting them to improve the separability of the invariant feature space. The separability can also be improved by constructing weak $G$-commutative maps acting on the feature space \citep{SM_Existence}. In general, the group average is closely related to the group convolution (cf. Eq. \ref{eq:GConvCont}). The special case where the function is a locally applied learnable kernel followed by average pooling is equivalent to a Group Lifting Convolution followed by Global Average Pooling.

\cite{SM_Algos} expands Invariant Integration to non-compact groups and continuous signals. Generally, combining invariance to multiple compact subgroups induces invariance to more general groups. For example, features invariant to the general linear group $GL(n,\mathbb{C})$ can be built by a quotient of homogeneous features invariant to the special unitary group $SU(n, \mathbb{C})$. Continuous signals within a Hilbert space are decomposed using a basis and an appropriate inner product to construct stable subsets. Subsequently, invariants to these subsets can be determined.

\cite{SM_Gray} applies the derived methods to 2D transformations and rotations on gray-scale images. The calculated invariant features are used to classify objects in an image with nearest neighbor classifiers. Twelve different monomials are processed via the group average to obtain invariance to global rotations and translations. In practice however, there might not only exist global transformations but also multiple local ones. In this case, the group-averaged monomials are not fully invariant, but vary only slowly. This also holds for small overlaps and articulated objects. Multiple objects within the same image are additive, i.e. their invariant feature values add, as long as they do not overlap. While the obtained features are invariant, they are not optimal for the specific application. Additional desirable properties, e.g. separability or robustness to distortions, need to be included by designing additional maps before applying invariant integration.

\cite{Condurache} use a Fourier transformations followed by invariant integration to achieve invariance to multiple transformations. First, they compute Fourier descriptors of human contours to obtain features invariant to color changes, starting point, rotation and translation. Moreover, they apply invariant integration with monomials in the Fourier domain. The obtained rotation-invariance in the Fourier space corresponds to invariance w.r.t. anthropomorphic changes, i.e. different size and build among humans, in signal space. In total, this approach can be seen as using a chain of invariant transformations to achieve an induced invariance in the input domain. Finally, a Support Vector Machine (SVM) processes the invariant features for human event detection in video scenes.

\cite{Rath} propose to use invariant integration with monomials in combination with equivariant group convolutions to compute a representation invariant to rotations and translations for image classification. The proposed Invariant Integration Layer's exponents $b_i$ can be optimized using the backpropagation algorithm which makes it a drop-in replacement for the spatial pooling layer usually utilized for the transition between the equivariant and invariant feature space in equivariant CNNs. By adding targeted model capacity, the data-efficiency of those networks is further improved. The monomials are selected with an iterative approach based on the least square error of a linear classifier following \cite{Muller3}.

\cite{Rath2} enhance the applicability of rotation-invariant integration within DNNs. First, they replace the iterative monomial selection with a pruning approach. Additionally, the monomials within invariant integration are replaced entirely by well-established functions such as self-attention and a weighted sum. Rotation-invariant integration using a weighted sum achieves a similar performance as the monomial-counterpart while streamlining the training procedure. Consequently, rotation-II can be applied to more complex datasets and models such as Wide-ResNets. 

\cite{Rath3} apply Invariant Integration in combination with group-equivariant convolutions beyond rotations. First, they include flips to obtain invariance to E(2). Then, an expansion that achieves scale-invariance is proposed which relies on computing the derivative between invariant integrals over homogeneous functions of the same order. Finally, a multi-stream architecture is introduced that efficiently combines and automatically selects important invariants for the task at hand. The architecture combines a scale-invariant, a E(2)-invariant and a standard convolutional stream in order to efficiently leverage invariants while allowing the flexibility to learn beyond the restricted filters in the invariant streams.

The group average (Equation \ref{eq:groupAverage}) has also been used beyond the invariant integration framework. TI-Pooling \citep{Laptev} can be interpreted as a case of invariant integration, where a CNN is used for the function $f$. The input is transformed with all transformation elements the network should be invariant to. One forward-pass per transformation element is computed using shared weights and max-pooling is applied among all responses. 
While this procedure is a straight-forward way to obtain transformation-invariance, it is computationally expensive since - due to its \textit{brute-force} nature - $\|G\|$ forward-passes need to be computed per input, also at test time. This is especially problematic for bigger transformation groups.

\cite{Puny} provide a method to solve the group average for larger, intractable groups by integrating over a subset called frame. They apply their method to classification for motion-invariant point clouds and graph DNNs integrating over the whole DNN as function $f$. 

\cite{Elesedy} use the Group Average to quantify the generalization benefits of in- and equivariant networks compared to their non-equivariant counterparts. They provide the first provable non-zero improvement for the generalization capability of in- or equivariant models, if the embedded equivariance is present in the target distribution. Finally, they provide a regularization term based on their results that can be used to enforce the desired invariance.

In summary, Invariant Integration is a method to construct a complete feature space w.r.t. a transformation. Comparably to the scattering transformation, most proposals use it to construct a feature space for classifiers such as a SVM, while novel approaches combine it with equivariant CNNs during the transfer from equi- to invariance. By adding targeted model capacity during the transition, the sample complexity of equivariant CNNs can be further improved.

\subsection{Other Methods}
In this section, we introduce approaches which do not fit into one of the previous subcategories.
\cite{GensDeepSymmetry} form feature maps over arbitrary symmetry groups via transformed kernels. A kernel-based interpolation scheme is used to find all transformed points in the lower layer contributing to a point of the symmetry feature map. The same kernel is used at every point of the symmetry feature map which leads to weight-tying among transformations. To reduce the computational complexity especially for higher-order transformations, a sub-sampling of the $N$ most important points in a neighborhood, determined by Gauss-Newton optimization, is performed.

%
%

\section{Training Procedure}\label{sec:Approximated}
We now present work that achieves in- or equivariance by adapting the training procedure of DNNs. In contrast to the architecture restrictions presented in section \ref{sec:Guaranteed}, the obtained equivariance properties are approximately learned rather than mathematically guaranteed. 

\subsection{Data Augmentation}
Data Augmentation is a general, straightforward training method to approximate in- or equivariance properties with a DNN. Formally, the training samples $\mathbf{x}$ and targets $\mathbf{y}$ are transformed with elements from a set of possible transformations such that the target encodes the desired transformation behavior. For example, the target remains unchanged if invariance is desired. During training, the specific transformation parameters are usually randomly sampled for each training sample. Thereby, the DNN generalizes among the defined transformation set by learning from transformed sample-target pairs. 

Data Augmentation often significantly improves the performance and robustness of DNNs and is easy to implement, as only the input and desired output transformations need to be known. The transformations are not restricted to groups, which enables an easy generalization to complicated augmentations such as occlusions, inpainting, lighting changes or even adding or removing objects from a scene.

On the other hand, DNNs trained with Data Augmentation are not guaranteed to exhibit the desired in- or equivariance but need to learn it during training. Therefore, additional model capacity is required, e.g. in early CNN layers, where multiple transformed versions of the same filter need to be learned. The desired symmetry properties are also not guaranteed per layer, but only globally for the entire model. During train time, more computation is required since each training sample needs to be randomly transformed. Moreover, the exact transformation set used for the augmentations can be hard to optimize -- mis-specified hyperparameters can even lead to a performance loss. Finally, defining the augmentation set still requires prior knowledge about the desired in- or equivariance properties.

A closely related approach to increase the robustness of DNNs is \textit{Test Time Augmentation}. Multiple transformed versions of the input are usually processed by the same network to enable Monte Carlo Sampling. The responses are then aggregated via averaging or a Bayesian Neural Network. While this approach increases the robustness, multiple parallel computations increase the memory and run-time required for inference.

As a final remark, we would like to emphasize that Data Augmentation methods can easily be used in conjunction with the earlier presented approaches that guarantee equivariance. This allows to combine mathematically guaranteed equivariance to well-defined transformation groups with robustness to more complicated transformations that cannot be modeled as groups.
  
\subsection{Symmetry Regularization}
A possibility to improve the transformation robustness learned with data augmentation is to add a regularization term that enforces the desired in- or equivariance properties on the latent representations or outputs. Similarly to Data Augmentation, the regularization term can be applied to transformations beyond groups, as long as the desired effect on the learned representations or output is known.

\cite{Coors2018VISAPP} process two different transformed versions of the same input and minimize their similarity loss based on the Kullback-Leibler divergence. The loss is computed at each latent representation of the DNN for each element and can be used for semi-supervised learning. Especially in the small sample regime, this approach improves the performance compared to plain data augmentation. Nevertheless, it does not reach the performance of guaranteed equivariant models on Rotated-MNIST.

\cite{RegularizedInvariance} model the worst-case transformations a DNN needs to be robust to as spatial adversarial attacks. Enforcing robustness to those attacks leads to a regularization term that is used in combination with data augmentation to encourage learning constant feature values for all transformed versions of the input. Consequently, invariance to transformation sets is guaranteed. This is a weaker requirement than invariance to transformation groups that allows to impose the desired properties on subsets not forming a group, e.g. rotations in $[-90^\circ, 90^\circ]$. In contrary to usual adversarial defense regularization, the proposed method does not include a trade-off between accuracy and adversarial robustness, but is able to improve the overall performance of the DNN.

\cite{shakerinava2022structuring} learn equivariant representations with a non-generative approach. Instead of restricting the architecture or the individual layers, the equivariant representations are characterized via the \textit{defining action} that should be preserved under input transformations, e.g. distance for the Euclidean groups. A group-specific regularization term  enforces the geometric invariants in the latent space via simple actions of the group in latent space. This approach can even be applied for non-linear and unknown actions in the input domain. Furthermore, the learned representations can be disentangled via group decomposition.

\cite{Bardes} propose a triple objective that is used for self-supervised learning. The desired embedding is learned via a term preventing embedding collapse and a second term decorrelating the different dimensions of the embedding. Finally, a \textit{invariance criterion} is applied that is calculated via the mean-squared Euclidean distance between the learned embeddings of a pair of transformed inputs.

\subsection{Training Scheme}
Another possibility to approximate in- or equivariance are adapted training schemes. \cite{Feige} uses a Variational Autoencoder (VAE) to learn an invariant class representation and equivariant information about the transformation needed to obtain the specific sample from the canonical group sample. The in- and equivariant representations are learned via an adapted training scheme. During reconstruction, the decoder only gets the invariant latent variable from samples of the same class, but not the sample itself. Contrarily, the equivariant information is sample-specific and is used to adapt the reconstructed class sample to the specific instance using a smooth transformation. Hence, the equivariant representation is able to encode arbitrary smooth transformations.

\subsection{Learned Feature Transformations}
A transformation invariant feature space can also be obtained by estimating the transformations acting on the input and re-transforming the signal to its \textit{canonical} form. This procedure can also be applied to the feature space of DNNs -- not only on the input. Hence, we refer to this approach as \textbf{learned feature transformations}.

\cite{HenriquesWarpedConv} propose to warp the input of a CNN using the exponential map of arbitrary transformation groups. Transformations of the input result in translations of the warped image map. Hence, a standard convolution can be applied in the warped space to achieve equivariance - decreasing computational requirements compared to group convolutions. This approach is applied to three transformations: scale and aspect ratio, scale and rotation, and perspective 3D rotations. However, in their implementation, the arbitrary groups need to be Abelian and only have two parameters in order to avoid more complex computations for the exponential map. 

\textit{Spatial Transformer Networks} (STNs, \cite{STN}) implicitly learn invariance to affine transformations including translation, rotation and scale without additional supervision. A three-step approach is used to estimate and perform transformations of the input or feature spaces of CNNs to obtain invariant representations: A localization network estimates the transformation parameters of an affine transformation, which is used to calculate a parameterized sampling grid. A differentiable image sampling, e.g. bi-linear interpolation, is then used to reversely transform the input.
This allows to use backpropagation to train both the convolution operator as well as the localization network at once.

\cite{PolarTN} enhance STNs by using a polar coordinate transformation to learn features for object classification that are invariant to translations and equivariant to rotations and scale. A CNN predicts the object center which is used as the origin for the log-polar transformation of the input. Effectively, the transformed representation is invariant w.r.t. the predicted object location. Additionally, rotations and scales in the regular image domain appear as shifts in polar space because they are the \textit{canonical coordinates} of those transformations. Consequently, a standard CNN can be used to process the log-polar representation for rotation- and scale-equivariant classification. Since the log-polar transformation is differentiable, the entire \textit{Polar Transformer Network} architecture including the polar origin predictor can be optimized end-to-end.

\cite{ETN} further generalize this method towards arbitrary continuous transformation groups and call their method \textit{Equivariant Transformer Networks} (ETNs). Similar to the procedure in STNs, a separate network is trained to estimate how the input is transformed. The estimated parameters are used to re-transform the input back to its \textit{canonical} form. However, in contrast to STNs, the input is transformed to the canonical coordinate system corresponding to the desired symmetry transformation. Thereby, the inverse transformation parameters are estimated in an equivariant manner using standard convolutions. In comparison to Polar Transformer Networks, ETNs only transform their input to canonical coordinates to estimate the transformation parameters. The feature extraction and classification are performed in the regular Euclidean domain.

While learned feature transformations facilitate learning invariant representations, they do not incorporate geometrical prior knowledge in a guaranteed way. In contrast, invariance is learned in a task-specific manner without any additional supervision. Nevertheless, \cite{ETN} and \cite{PolarTN} combine equivariant convolutions and learned feature transformations to achieve learned invariance with additional equivariance guarantees. Additionally, similar to Data Augmentation, the learned transformations are more flexible than provably in- or equivariant layers, since they are not restricted to groups.

In summary, the training procedure of DNNs can be adapted such that in- or equivariance properties are learned using data augmentation, symmetry regularization, adaptive training schemes or learned feature transformations. While these approaches are more flexible than architecture restrictions and not limited to group transformations, they approximate the desired properties rather than guaranteeing them. Hence, they achieve worse performance when exact in- or equivariance is desired. Additionally, the adapted training procedures can be easily combined with restricted architectures.  

\section{Discovering In- \& Equivariance from Data}\label{sec:Discovered}
Multiple approaches try to learn the desired in- or equivariance directly from data rather than guaranteeing it via fixed, pre-determined architecture restrictions or training procedures. On the one hand, this allows to discover symmetries automatically, which is helpful if the required prior knowledge is not available. On the other hand, this approach does not strictly leverage prior knowledge -- and will at most achieve the guaranteed equivariance performance, if the built-in symmetries are present in the data.

\cite{vanDerWilk} learn invariances for Gaussian process models in a Bayesian manner. Therefore, the data transformations are modeled as priors. A variational lower bound for the models' kernels is computed by sampling points from a distribution that describes the required invariance. The transformation parameters can then be learned by optimizing the marginal likelihood bound via backpropagation. 

\cite{LearningInvariances} propose Augerino, a method to learn distributions over the parameters of affine data augmentations. During training, the reparametrization trick is applied s.t. gradients w.r.t. augmentation parameters of Lie groups can be computed from a single augmented sample via backpropagation. A regularization term is applied that facilitates the selection of non-zero augmentations. During inference, test time augmentation is employed, i.e., multiple augmented samples are processed and averaged to achieve the final prediction. While this allows to learn augmentations directly from the training data, it increases the inference time by the number of samples required for the final prediction.  

\cite{DifferentiableAugmentationLayers} expand Augerino  beyond Lie groups using hierarchical, differentiable data augmentation layers that are embedded into the network. Additionally, the model learns the weights of a weighted average rather than using the fixed average, which allows to learn the importance of each transformation in addition to its range. The proposed expansions both increase the performance of Augerino, while allowing to apply it beyond Lie Groups and computer vision problems.

\cite{ImmerInvarianceLearning} optimize the hyper-parameters of data augmentations via their Marginal Likelihood which corresponds to Bayesian Model Selection. They use a differentiable Kronecker-based Laplace approximation to allow computation for large-scale neural networks as well as gradient-based optimization directly from training data. 

\cite{Dehmamy} introduce a Lie algebra based convolutional network. The Lie algebra basis provides equivariance to continuous groups without discretization or summing over irreducible representations. Additionally, the basis can be learned from data, enabling to automatically discover symmetries instead of embedding them a-priori. Lie algebra CNNs are able to represent G-Convs on any compact group G as a special case. 

\cite{VanDerOuderaa} learn the parameters of affine invariance groups via the evidence lower bound of the marginal likelihood using Monte Carlo sampling. However, this approach is only applied to single-layer DNNs using continuous transformations. Expansions towards discrete groups are hard to calculate, while for deeper networks, the value of the lower bound remains to be proven.

\cite{ResidualPathwayPriors} soften built-in architectural equivariance constraints by adding an unconstrained residual path to the model. By selecting appropriate priors, the network is still biased towards the equivariant solution, but can benefit from the additional model capacity, when the desired symmetry only holds approximately or is misspecified.

\cite{LearningPartialEquivariance} introduce partial group-convolutions that achieve approximate equivariance to subsets of the equivariance group $G$. Equivariance to subsets is achieved by using the Monte Carlo approximation of the group convolution and sampling the group elements from learnable probability distributions. For discrete groups, those distributions are learned via the Gumbel-Softmax trick. For continuous groups, the reparameterization trick is applied to learn a uniform distribution over the Lie algebra. Partial G-Convs are able to learn the desired equivariance subset per layer, allowing varying equivariance constraints throughout the network.

\cite{RelaxingEquivarianceConstraints} propose a generalization of the group-equivariant convolution that allows to relax strict symmetry constraints. A non-stationary kernel that also depends on the absolute input group-element can interpolate between a non-equivariant linear product, a strict equivariant convolution and a strict invariant map. The symmetry constraint is tunable in the Fourier domain and can be directly learned from the training data. Thus, this approach allows to learn the desired equivariance and prevents too strict equivariance constraints while allowing to incorporate the prior knowledge about symmetries. 

\cite{MetaLearningSymmetries} propose to meta-learn parameter-sharing patterns from data in order to achieve guaranteed equivariance to any finite symmetry group. However, this approach is constrained to symmetries shared among multiple tasks due to the meta-learning setup.

\cite{MiyatoEquivarianceSequences} train models to automatically discover in- or equivariances from time series data that exhibits stationarity properties. Therefore, the model is trained to predict future observation where the transition of the latent variables are restricted to be linear. The underlying symmetries of the dataset can then be discovered by block-diagonalizing the linear transitions, where each block represents a certain factor of variation.

To summarize, the desired symmetries of a task can be directly learned from training, if the required prior knowledge is not fully available. Different approaches learn the transformation parameters of Data Augmentations or parameters of (relaxed) group-equivariant convolutions directly from the training data. Moreover, special training schemes can be used to model transformation distributions, e.g. via Monte Carlo sampling. While automatically learning symmetries is beneficial when prior knowledge is unavailable, misspecified, or hard to determine, e.g. for latent representations, it is unable to outperform DNNs with correctly specified, built-in in- or equivariance properties. 

\section{Measuring Equivariance}\label{sec:Measured}
Another recent field investigates, to what extent DNNs are in- or equivariant to the desired symmetry transformations. \cite{LieDerivativeLearnedEquivariance} measure the equivariance error of benchmark vision architectures using the Local Equivariance Error that is based on the Lie derivative. Their experiments show a strong correlation between a lower equivariance error and a better task performance. Additionally, given enough data and the correct training procedure with data augmentations, non-equivariant models such as Vision Transformers can achieve a lower equivariance error than equivariant architectures. For example, a pre-trained transformer performs only slightly worse than the best rotation-equivariant network on Rotated MNIST. By computing the layer-wise equivariance error, the authors identify point-wise non-linearities to be responsible for breaking equivariance guarantees due to aliasing effects, which allows training procedures to achieve the same equivariance properties than specifically designed architectures.

\cite{kvinge2022in} propose to directly measure the in- and equivariance properties of DNNs with a family of metrics called G-empirical equivariance deviation. It is defined as the distance between the model's output under input transformations and the transformed expected output - averaged over a given dataset and a transformation group. Thereby, it measures the extent a DNN fails to be in- or equivariant. The proposed measure is applied to gain insights about invariant models obtained via data augmentation or G-equivariant layers. Most importantly, models trained with augmentation do not achieve layer-wise, but only global invariance. Moreover, the invariance properties of G-convolutional models generalize better to out-of-distribution setups.

\cite{LieGG} extract the symmetries learned by DNNs by computing the Lie group generators depending on the specific training data and the model derivative. Thereby, the specific invariances and their degree can be quantified without specifying a set of transformations beforehand, which allows to investigate the invariance properties of different DNNs. In the presented experiments that are restricted to small-scale MLPs and datasets with well-controlled factors of variation, models with more parameters and gradual fine-tuning outperform their counterparts in terms of equivariance guarantees.

\section{Datasets and Benchmarks}
\begin{figure}
	\centering
	\begin{subfigure}{0.07\linewidth}
		\includegraphics[width=\linewidth]{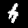}
	\end{subfigure}
	\begin{subfigure}{0.07\linewidth}
		\includegraphics[width=\linewidth]{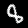}
	\end{subfigure}
	\begin{subfigure}{0.07\linewidth}
		\includegraphics[width=\linewidth]{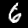}
	\end{subfigure}
	\begin{subfigure}{0.07\linewidth}
		\includegraphics[width=\linewidth]{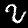}
	\end{subfigure}
	
	\vspace{2pt}
	\begin{subfigure}{0.07\linewidth}
		\includegraphics[width=\linewidth]{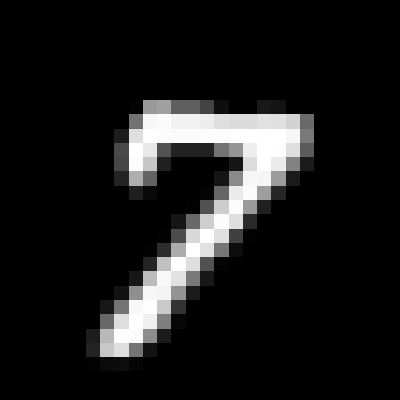}
	\end{subfigure}
	\begin{subfigure}{0.07\linewidth}
		\includegraphics[width=\linewidth]{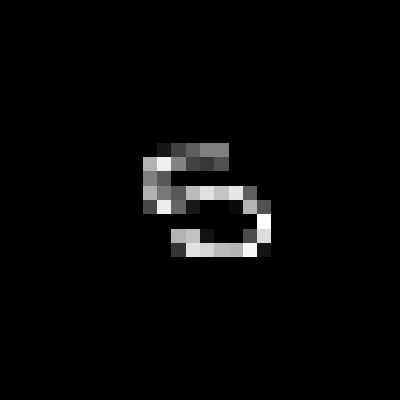}
	\end{subfigure}
	\begin{subfigure}{0.07\linewidth}
		\includegraphics[width=\linewidth]{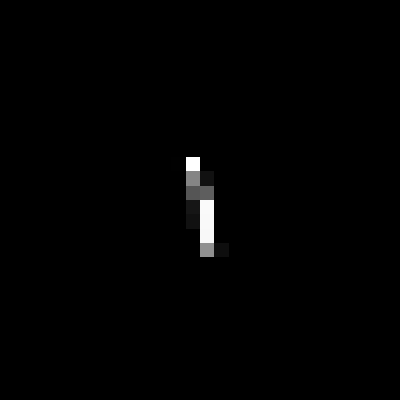}
	\end{subfigure}
	\begin{subfigure}{0.07\linewidth}
		\includegraphics[width=\linewidth]{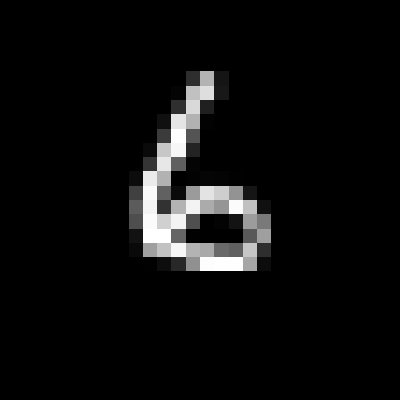}
	\end{subfigure}
	
	\vspace{2pt}
	\begin{subfigure}{0.07\linewidth}
		\includegraphics[width=\linewidth]{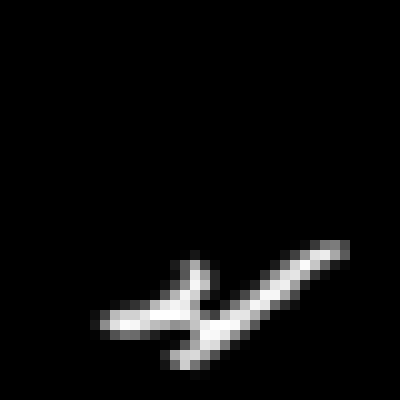}
	\end{subfigure}
	\begin{subfigure}{0.07\linewidth}
		\includegraphics[width=\linewidth]{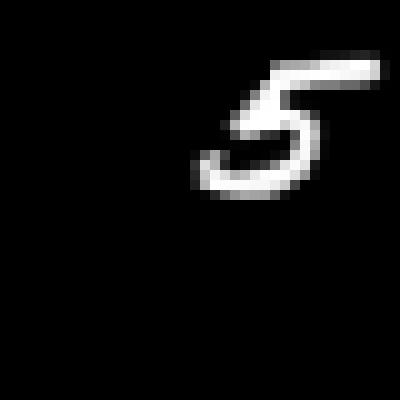}
	\end{subfigure}
	\begin{subfigure}{0.07\linewidth}
		\includegraphics[width=\linewidth]{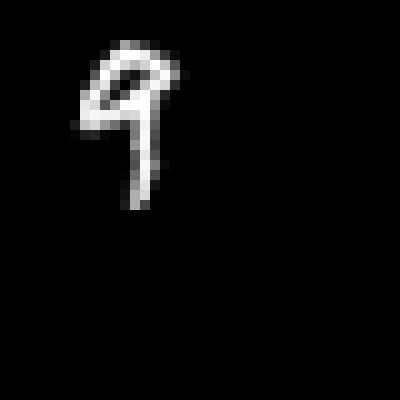}
	\end{subfigure}
	\begin{subfigure}{0.07\linewidth}
		\includegraphics[width=\linewidth]{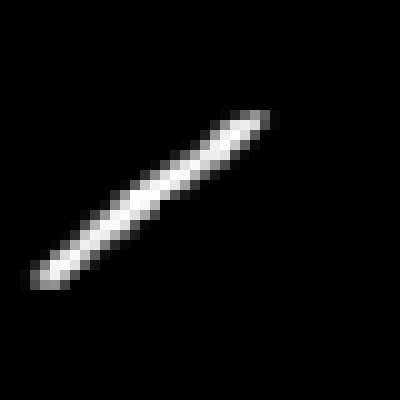}
	\end{subfigure}
	\caption{Examples from rotated MNIST (top), MNIST-Scale (middle) and affNIST (bottom).}
	\label{fig:MNIST}
\end{figure}
\begin{table*}[t]
	\centering
	\small
	\begin{tabular}{lc}
		\toprule
		Method & Mean Test Error [$\%$] \\ 
		\midrule
		Conventional CNN & 5.03 \\
		\cite{Scattering} & 4.4 \\
		\cite{LenssenGroupCapsules} & 2.6\\
		\cite{PolarTN} & 1.83 \\
		\cite{HarmonicNetworks} & 1.69 \\
		\cite{Laptev} & 1.2 \\
		\cite{MarcosRot} & 1.09 \\
		\cite{Rath} & 0.72 \\
		\cite{SFCNN} & 0.71\\
		\cite{Rath2} & 0.69 \\
		\cite{E2STCNNs} & 0.68 \\
		\bottomrule
	\end{tabular}
	\caption{Mean test error (MTE) $[\%]$ on rotated MNIST using 12k training samples trained with rotation augmentations.}
	\label{tab:RotMNIST}
\end{table*}
\begin{table*}[t]
	\centering
	\small
	\begin{tabular}{lcc}
		\toprule
		Method & \multicolumn{2}{c}{MTE[\%]} \\ \cmidrule{2-3} 
		& \multicolumn{2}{c}{Augmentation}\\ 
		& \ding{55} & \checkmark \\
		\midrule
		CNN$^\dagger$ & 2.02 & 1.60 \\
		\cite{XuScale}$^\dagger$ & 2.02 & 1.59 \\
		\cite{LocalScaleInvariance}$^\dagger$ & 1.82 & 1.59 \\
		\cite{MarcosScale}$^\dagger$ & 1.87 & 1.62 \\
		\cite{Worrall19}$^\dagger$ & 1.92 & 1.57 \\
		\cite{GhoshScale}$^\dagger$ & 1.84 & 1.76 \\
		\cite{Sosnovik} & 1.68 & 1.42 \\
		\cite{DISCO} & 1.52 & 1.35 \\
		\cite{Rath3} & - & 1.30 \\
		\bottomrule
	\end{tabular}
	\caption{Mean test error (MTE) $[\%]$ on MNIST-scale using 10k training samples and bi-linear interpolation to images of size 56x56. $^\dagger$: reported by \cite{Sosnovik} who trained all models in similar training settings rather than original reported result.}
	\label{tab:ScaledMNIST}
\end{table*}
\begin{table*}[t]
	\small
	\centering
	\begin{tabular}{lc}
		\toprule
		Method & Test Error [$\%$] \\ 
		\midrule
		Conventional CNN \citep{DynamicRoutingCapsules} & 34.0 \\
		\cite{DynamicRoutingCapsules} & 21.0 \\
		Conventional CNN \citep{ETN} & 12.7 \\
		\cite{LenssenGroupCapsules} & 10.9 \\
		\cite{Capsules} & 6.9 \\
		\cite{VenkDeepEquivariantCapsules} & 2.99 \\
		\cite{ETN} & 1.7 \\
		\bottomrule
		\end{tabular}
	\caption{Test error $[\%]$ on the affNIST dataset.}
	\label{tab:AffNist}
\end{table*}
\begin{table*}[t]
	\small
	\centering
	\begin{tabular}{lcc}
		\toprule
		Method & Invariance & Test Error [$\%$] \\ 
		\midrule
		\cite{WideResNet} & $\mathbb{Z}^2$ & 11.48 \\
		\cite{Rath2} & SE(2) & 10.09 \\
		\cite{E2STCNNs} & E(2) & 9.80 \\
		\cite{Sosnovik} & $\mathbb{R}^2 \rtimes S$ & 8.51 \\
		\cite{DISCO} & $\mathbb{R}^2 \rtimes S$ & 8.07 \\
		\cite{Rath3} & E(2) \& $\mathbb{R}^2 \rtimes S$ \&  $\mathbb{Z}^2$  & 6.46 \\ 
		\bottomrule
	\end{tabular}
	\caption{Test error $[\%]$ on STL-10 using Wide ResNet 16-8 \citep{WideResNet} with constant number of parameters as baseline architecture. All models have roughly the same number of parameters.}
	\label{tab:STL}
\end{table*}
Many algorithms presented in this survey evaluate their performance on artificially transformed versions of MNIST \citep{MNIST}, a well-known dataset for hand-written digit recognition, to prove the incorporated in- or equivariance properties. We summarize the reported performances of different in- or equivariant algorithms on transformed MNIST datasets in this section to allow for an easy comparison.

The rotated MNIST dataset \citep{RotMNIST} contains digits that are randomly rotated by angles $\phi \in (0, 360)^\circ$. It consists of a training set with $10000$, a validation set with $2000$ and a test set with $50000$ samples. Examples of the dataset are shown in Figure \ref{fig:MNIST} (top).
Due to the relatively small amount of training samples, the variability of all possible rotations is not captured in the training dataset. Hence, DNNs that guarantee in- or equivariance to rotations achieve comparably better results. The reported results of various proposals are shown in Table \ref{tab:RotMNIST}.

MNIST-Scale \citep{MNIST-Scale} consists of digits which are randomly scaled by uniformly sampled factors $s \in (0.3, 1)$. The size of the training, validation and test set is the same as for rotated MNIST. Examples are depicted in Figure \ref{fig:MNIST} (middle) and the reported results are shown in Table \ref{tab:ScaledMNIST}. 

Finally, the affNIST dataset \citep{affNIST} is used to test the robustness of algorithms, mainly capsule networks, to unseen affine transformations. Therefore, the networks are trained on randomly translated digits from the MNIST dataset and tested on digits which are randomly transformed with small affine transformations. Examples are shown in Figure \ref{fig:MNIST} (bottom) while the reported results are listed in Table \ref{tab:AffNist}.

Additionally, most authors also apply their approach to a real world dataset where the incorporated prior knowledge often allows to outperform conventional algorithms, especially in the limited data domain. One commonly used dataset is STL-10 \citep{STL10}, a subset of ImageNet tailored to the research community for self-supervised learning and learning from limited datasets. It consists of 5,000 training and 8,000 validation images containing ten different classes. Reported results are shown in Table \ref{tab:STL}. We refrain from listing further datasets since their uniqueness and variety prevents us from comparing multiple algorithms. 

\section{Towards Geometrical Prior Knowledge for Autonomous Driving}
Systems for automated driving need to solve a variety of problems that benefit from deep learning approaches. This includes processing and fusing signals from different sensor modalities, perceiving the environment, i.e. reliably detecting and locating objects in a driving scene, understanding and assessing the environment, localizing the ego vehicle and finally, planning and executing driving maneuvers. Especially for the perception task, deep learning approaches have been established to provide state-of-the-art results on several benchmarks (e.g. \citealt{KITTI3D, nuScenes}). In addition, reinforcement or imitation learning approaches can be used to tackle the autonomous driving task in a more coarse-grained, end-to-end manner \citep{LearningFromAllVehicles, ImitationLearningAD}. Nowadays, DNNs are used within both prototype and production vehicles that already drive on real streets at least partially automated. We refer the reader to \cite{SurveyAutonomous} for a concise overview over datasets, methods and challenges concerning deep learning and autonomous driving.

Nevertheless, there remain some open problems which need to be solved before DNNs can safely be used within self-driving cars. Foremost, collecting and labelling the data needed for supervised learning is both time-consuming and cost-expensive. Particularly, labelling three-dimensional objects in driving scenes is a non-trivial task. Furthermore, DNNs are hard to evaluate due to their black box behavior which makes it hard to predict outcomes given data-points the network has not been trained on. This is delicate since self-driving cars must operate safely even in unexpected circumstances, e.g. a car behaving in a way that was not included in the training data.

We argue that incorporating geometrical prior knowledge to deep learning systems for autonomous driving helps to mitigate both presented problems. On the one hand, the data-efficiency of DNNs is increased which means that less data needs to be collected, stored and labelled to achieve the same performance level. On the other hand, in- or equivariance guarantees promote the interpretability of DNNs. 

\paragraph{Equivariant 3D Object Detection}
As an example, we show how 3D object detection can benefit from in- or equivariance to various transformations. In this section, we list several transformation groups that affect the output of an 3D object detector in a predictable way. 
Ideally, a 3D object detector operating on images should be stable to shape deformation, varying object size (i.e. scale) or color and illumination changes --  and equivariant to translation and rotation of the object to be detected since the pose information is needed to predict the exact location. Moreover, DNNs processing 3D data as point clouds or graphs must be invariant to permutations or varying point cloud densities and equivariant to 3D translations and rotations. Ideally, the detection should also be robust to more complicated transformations, e.g. (partial) occlusions. 

Object detection can be divided in several sub-tasks: Recognition, object type classification and estimation of the pose (relative position, orientation and size). While recognition and classification build invariants, it is important to preserve information about position, orientation and size to be able to estimate them. Therefore, equivariant outputs are more appropriate in this case. As presented in this survey, the desired properties to well-defined transformations could be incorporated via architecture restrictions. Automatic symmetry discovery methods even enable an improved sample complexity, if the exact invariance set is not entirely known a-priori. Moreover, more complex transformations such as occlusions could be handled via adapted training procedures, e.g. data augmentation.

\section{Conclusion}
In this section, we first summarize the various methods presented and their different advantages and disadvantages. Afterwards, we draw a conclusion and give a short outlook on possible future work. 

\textbf{Restricting DNN architectures} by using fixed filters such as the \textbf{scattering transformation} provides a provable way to obtain in- or equivariant representations in neural networks and reduce the space of learnable parameters. However, those restrictions are often too strict and prevent the network from learning informative representations, especially when all filter coefficients are fixed rather than learnable. 

Enforcing equivariance properties using \textbf{group convolutions} provides a mathematically guaranteed way to incorporate geometrical prior knowledge to DNNs. Since the coefficients are still learnable, G-CNNs achieve state-of-the-art results in a variety of tasks benefiting from in- or equivariance properties. Up to now, group convolutions are restricted to fairly simple and often finite transformation groups and suffer from a computational overhead. Moreover, important symmetries that occur in many tasks are not expressible using the concept of groups, e.g. viewpoint changes in images. Incorporating in- or equivariance to multiple transformation groups at once is also an area which needs further investigation. The concept of achieving guaranteed equivariance via a generalized equivariant operation has further been expanded to \textbf{non-linear equivariant maps} such as self-attention. In general, these approaches achieve the same benefit of an increased sample complexity, at the downside of a higher computational complexity.

\textbf{Capsules} disentangle learned object representations into invariant object information and equivariant pose parameters. This includes sophisticated transformations such as viewpoint-changes or lighting which can not be modeled as groups. However, capsules do not guarantee the desired properties. Adaptations leveraging group symmetry exist but either suffer from a reduced transformation efficiency or restrictions regarding the equivariant pose information. Capsule networks are non-straightforward to train since they rely on a carefully-designed routing algorithm to assign activations to other layers as well as for backpropagation. Consequently, they often fail to reach the baseline performances provided by convolutional or self-attention based algorithms. 

\textbf{Invariant Integration} is a method to guarantee invariance while increasing the separability of distinct patterns. It can thus be used to enhance the transfer from equi- to invariant representations via additional, targeted model complexity. However, it suffers from computational complexity, is restricted to group transformations disposes the information about the symmetry group and, in the case of monomial-based invariant integration, the monomial parameters are hard to choose. Consequently, it is best used in combination with equivariant G-CNNs to improve the separability of learned invariant representations while exploiting the properties of equivariant learnable filters.

Other approaches adapt the \textbf{training procedure} rather than restricting the architecture to achieve in- or equivariance. \textbf{Data augmentation} is easy to implement and highly adaptable towards a variety of transformations. Furthermore, it can be used in combination with other approaches. On the downside, it does not provide any mathematical guarantees, augmenting the data with all possible transformations is computationally inefficient, the solution space is not restricted effectively and the equivariance is only learned for the network as a whole, not at each layer. Data augmentation can further be enhanced via \textbf{regularization} techniques that use additional losses to increase the robustness at the cost of requiring to process transformed pairs of inputs.

\textbf{Learned Feature Transformations} provide a way to learn invariance from data rather than defining and incorporating it manually. This is achieved by re-transforming the inputs to their canonical form via estimated parameters. Feature transformation layers are easy to include to existing neural network architectures and can also be applied to intermediate representations. The transformations are learned from data and not necessarily restricted to groups. Nonetheless, learned feature transformations lack mathematical guarantees and rely on estimating the transformation coefficients correctly.

\textbf{Automatic Symmetry Discovery} methods learn and incorporate in- or equivariance properties into the DNN directly from training data. This mitigates the problem of overly restrictive models due to misspecified invariances and allows to incorporate invariance when the required prior knowledge is not available. Nevertheless, it fails to outperform guaranteed methods, if the desired properties are known before training.

\textbf{Equivariance Measures} provide a way to investigate, if the desired properties are actually respected by the trained networks. In addition, it allows to identify weak spots of in- or equivariant DNNs, e.g. the Pooling layers that are detrimental to invariance guarantees due to aliasing effects. 

In this survey, we showed how utilizing prior knowledge can enhance state-of-the-art deep learning systems and used 3D object detection as an example. Foremost, defining problem-specific symmetry groups and introducing in- or equivariance to them can greatly improve the performance of neural networks, especially when the amount of training data is scarce. This has been proven for a broad variety of interesting applications from medical imaging to 3D shape recognition. Moreover, this approach improves the interpretability of neural network layers which is important for validation.

We expect future work to investigate further generalizations to more general input domains and symmetry groups as well as to more complex, non-invertible transformations, e.g. occlusions. Moreover, DNNs incorporating multiple symmetries at once could be of interest. Finally, improving the automatic inclusion of geometrical priors, e.g. via Neural Architecture Search, and providing a clear guideline on how and when to apply which in- or equivariances would be of great interest.

\section*{Statements and Declarations}
\subsection*{Conflicts of Interest}
All authors of this contribution are employed and funded by Robert Bosch GmbH and affiliated with the Institute for Signal Processing of the University of L\"ubeck.

\section*{Acknowledgments}
We would like to acknowledge our colleagues Julia Lust and Steffen Hagedorn as well as David W. Romero (Vrije Universiteit Amsterdam) for carefully reading the paper and their valuable remarks.

\bibliography{survey_bib}




\end{document}